\documentclass[10pt,journal,compsoc]{IEEEtran}
\ifCLASSOPTIONcompsoc
  \usepackage[nocompress]{cite}
\else
  \usepackage{cite}
\fi
\ifCLASSINFOpdf
\else
  Online Image Vectorizern (dvipsone, dvipdf, if not using dvips). graphicx
\fi

\usepackage{subfigure}
\usepackage{times}
\usepackage{epsfig}
\usepackage{hyperref}
\usepackage{graphicx}
\usepackage{amsmath}
\usepackage{amssymb}
\usepackage{multirow}
\usepackage{tabularx}
\usepackage{booktabs}

\usepackage{amsmath, bm}
\usepackage{algorithm}
\usepackage{xcolor,colortbl}
\usepackage{color, colortbl}
\usepackage{breakcites}
\usepackage{dsfont}
\usepackage{bbm}
\usepackage{makecell}
\usepackage{colortbl}
\usepackage{marvosym}

\usepackage{algorithmic}
\usepackage{array}
\usepackage{url}
\usepackage{verbatim}
\usepackage{cite}
\usepackage{url}

\newcommand{\DualUpdate}{Disocclusion Boundary Re-Injection}
\newcommand{\dualupdate}{disocclusion boundary re-injection}
\newcommand{\Vid}{\mathbf{X}}
\newcommand{\Frame}{X}
\newcommand{\VidDepth}{\mathbf{d}}
\newcommand{\FrameSub}{s}
\newcommand{\FrameSubMax}{S}
\newcommand{\Prompt}{c}
\newcommand{\LeftSub}{l}
\newcommand{\RightSub}{r}

\newcommand{\Generator}{\mathcal{G}}
\newcommand{\Noise}{\mathbf{\epsilon}}
\newcommand{\VidLatent}{\mathbf{z}}
\newcommand{\ImageMasks}{\mathbf{M}}

\newcommand{\LatentMasks}{\mathbf{m}}
\newcommand{\LatentMask}{m}
\newcommand{\ViewSub}{v}
\newcommand{\ViewSubMax}{V}
\newcommand{\NoiseScale}{\bar{\alpha}_t}
\newcommand{\Encoder}{\mathcal{E}}
\newcommand{\Decoder}{\mathcal{D}}
\newcommand*{\vertbar}{\rule[-1ex]{0.5pt}{2.5ex}}
\newcommand*{\horzbar}{\rule[.5ex]{2.5ex}{0.5pt}}

\definecolor{myblue}{RGB}{23,183,241}
\definecolor{mygray}{RGB}{230,230,230}

\begin{document}

\title{S{$^2$}VG: 3D Stereoscopic and Spatial Video Generation via Denoising Frame Matrix}

\author{Peng Dai, Feitong Tan, Qiangeng Xu, Yihua Huang, David Futschik, Ruofei Du, Sean Fanello, \\Yinda Zhang$^{\dag}$, Xiaojuan Qi$^{\dag}$
    \IEEEcompsocitemizethanks{
        \IEEEcompsocthanksitem Peng Dai, Yihua Huang, Xiaojuan Qi are with the Department of Electrical and Electronic Engineering at The University of Hong Kong, Hong Kong. 
        
        \IEEEcompsocthanksitem Feitong Tan, Qiangeng Xu,  David Futschik, Ruofei Du, Sean Fanello, Yinda Zhang are with Google, USA. 

        \IEEEcompsocthanksitem $\dag$ 
        indicates corresponding author
    
    }
}

\IEEEtitleabstractindextext{
\begin{abstract}
 While video generation models excel at producing high-quality monocular videos, generating 3D stereoscopic and spatial videos for immersive applications remains an underexplored challenge. 
We present a pose-free and training-free method that leverages an off-the-shelf monocular video generation model to produce immersive 3D videos. Our approach first warps the generated monocular video into pre-defined camera viewpoints using estimated depth information, then applies a novel \textit{frame matrix} inpainting framework. This framework utilizes the original video generation model to synthesize missing content across different viewpoints and timestamps, ensuring spatial and temporal consistency without requiring additional model fine-tuning. 
Moreover, we develop a \dualupdate~scheme that further improves the quality of video inpainting by alleviating the negative effects propagated from disoccluded areas in the latent space. 
The resulting multi-view videos are then adapted into stereoscopic pairs or optimized into 4D Gaussians for spatial video synthesis.  
We validate the efficacy of our proposed method by conducting experiments on videos from various generative models, such as Sora, Lumiere, WALT, and Zeroscope. The experiments demonstrate that our method has a significant improvement over previous methods. Project page at: \url{https://daipengwa.github.io/S-2VG_ProjectPage/}
\end{abstract}

\begin{IEEEkeywords}
    Video generation, Stereoscopic video, Spatial video, Inpainting, Diffusion model, Frame matrix, 4D Gaussian.
\end{IEEEkeywords}}

\maketitle

\IEEEdisplaynontitleabstractindextext

\IEEEpeerreviewmaketitle

\IEEEraisesectionheading{\section{Introduction}\label{sec:inrto}}

\begin{figure*}
    \centering
    \includegraphics[width=1.0\linewidth]{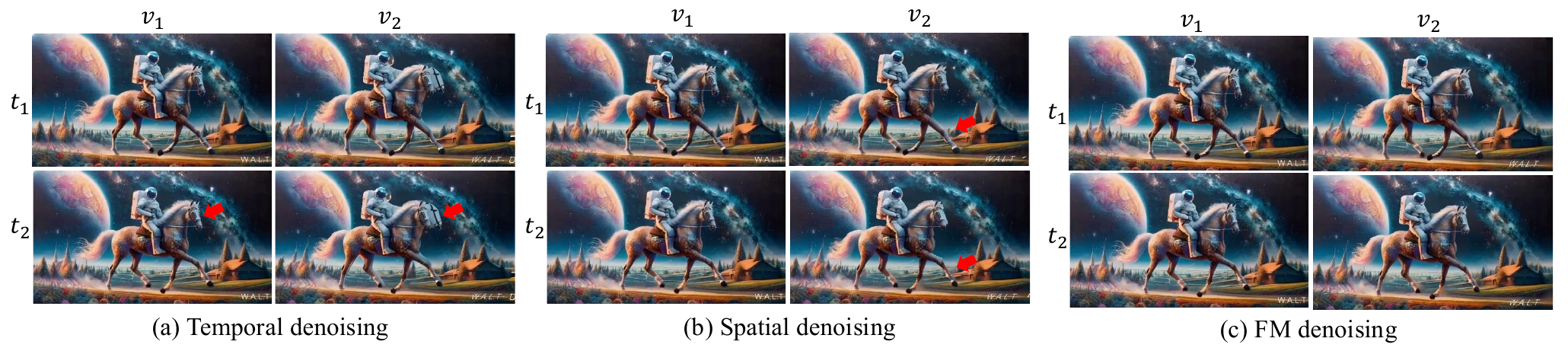}
    \vspace{-0.25in}
    \caption{{\textbf{Motivation of frame matrix denoising.} (a) When Denoising in the temporal direction, results are temporally consistent but lack semantic coherence. (b) When denoising in the spatial direction, results are semantically coherent but lack temporal consistency. (c) Frame matrix with alternative denoising enables leveraging priors from both directions to achieve outputs with enhanced consistency.} }
    \label{fig:mutal_denoising}
    \vspace{-0.1in}
\end{figure*}

\IEEEPARstart{T}{h}e rapid advancement of VR/AR technologies has created a growing demand for high-quality stereoscopic content capable of delivering truly immersive 3D experiences. Such content must maintain perfect 3D and semantic coherence between views while ensuring seamless temporal consistency across frames. Although monocular video generation has achieved remarkable progress, with current methods producing high-fidelity videos from complex text prompts~\cite{videoworldsimulators2024}, generating scene-level immersive 3D videos, encompassing both stereoscopic and spatial videos, remains an open challenge. This gap primarily arises from data limitations: while abundant training data exists for monocular video generation, comparable stereoscopic and multi-view video datasets remain scarce.

An emerging solution is to render immersive 3D videos from monocular videos using novel view synthesis~\cite{li2023dynibar,liu2023robust}. However, these reconstruction-based methods face two fundamental limitations. First, these reconstruction methods cannot generate plausible content for occluded regions absent from the input frames. Second, these approaches critically depend on highly accurate camera pose estimation, which is still challenging and unstable. 
Both structure-from-motion (SFM)~\cite{schoenberger2016sfm} and joint optimization~\cite{liu2023robust} easily lead to instability, particularly in dynamic scenes with subtle camera motions or when dynamic objects with temporal appearance variations dominate the visual content.


In this work, we first present a novel pose-free and training-free framework, for the sake of robustness and generalization capability, that uniquely leverages inference from an off-the-shelf video generation model~\cite{wang2023modelscope} to produce high-quality 3D stereoscopic videos. 
Our initial attempt follows a typical 2D-to-3D image uplifting methodology~\cite{hollein2023text2room} and extends it to the video domain. 
Specifically, we begin by generating a monocular video as the left view, which is then warped into the right view using estimated monocular depths~\cite{yang2024depth}, where we apply temporal smoothing to enhance the consistency of estimated depths and meticulously develop an outpainting and multi-plane warping pipeline to mitigate artifacts, such as cracks, isolated points, and partially observed content, caused by non-watertight and narrow-FOV RGB-D frames (see Fig.~\ref{fig:ab_data_preprocessing} and Fig.~\ref{fig:incomplete}).
Subsequently, we employ an off-the-shelf video generation model~\cite{wang2023modelscope} to synthesize natural right-view videos. This is achieved by iteratively adding and removing noise from warped frames, effectively inpainting disoccluded regions caused by the parallax of two views, drawing inspiration from diffusion-based image inpainting~\cite{avrahami2023blended}.

\begin{figure*}[t]
    \centering
    \includegraphics[width=0.98\linewidth]{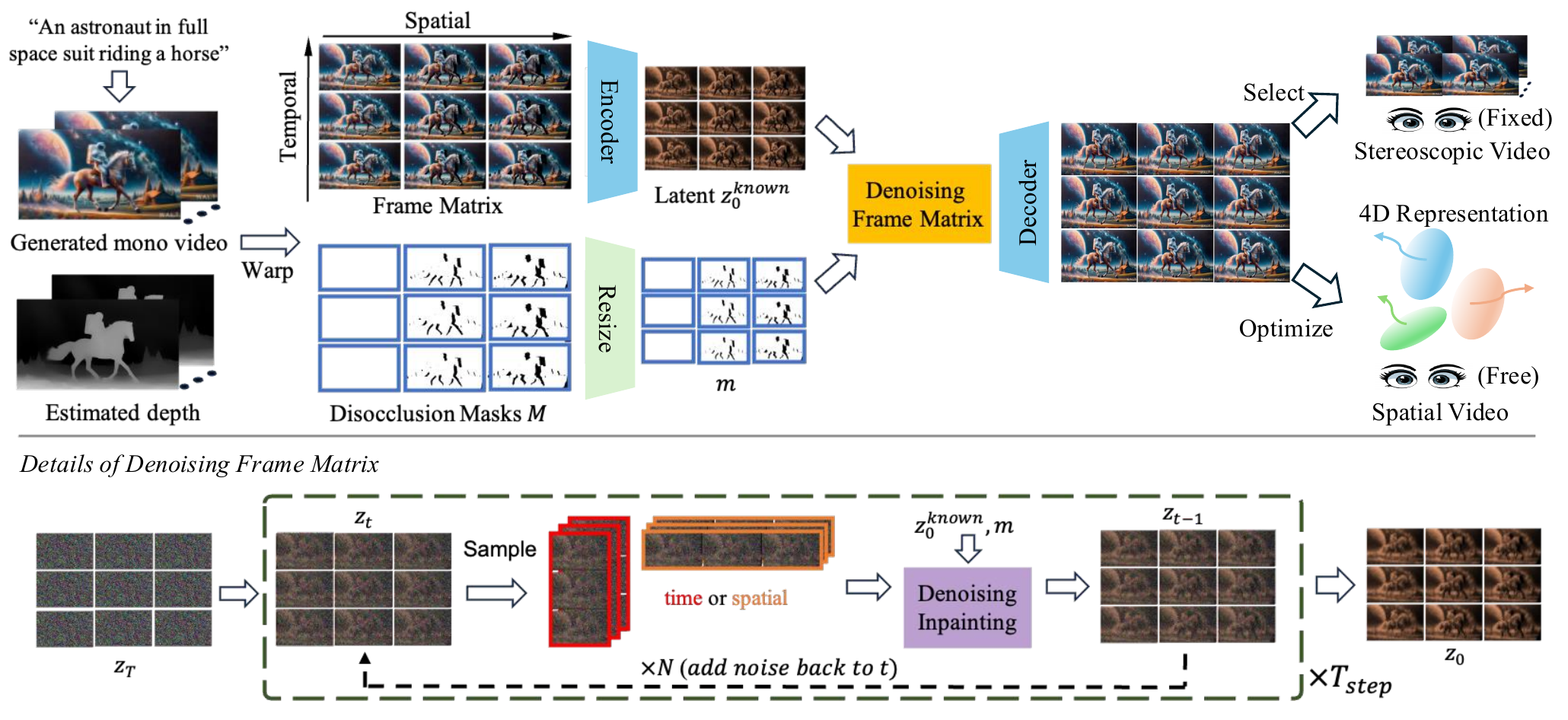}
    \vspace{-0.2in}
    \caption{\textbf{Overview} -- \textbf{Top:} Given a text prompt, our method first employs a video generation model to generate a monocular video, which is warped into pre-defined camera views using estimated depth to form a \textit{frame matrix} with disocclusion masks. Then, the disoccluded regions are inpainted by denoising the sampled frame sequences within the \textit{frame matrix}. After denoising, we decode the clean latent \textit{frame matrix} into RGB frames, where the 3D stereoscopic video is obtained via choosing the leftmost and rightmost columns. To generate spatial video, we optimize decoded frames into a 4D representation (e.g., 4D Gaussian) that supports continuous view changes. \textbf{Bottom:} Details of denoising \textit{frame matrix}. We initialize the latent matrix $\VidLatent_T$ as random noise maps, and alternatively denoise them in sampled directions. Specifically, we extend the resampling mechanism~\cite{karnewar2023holodiffusion,lugmayr2022repaint} to alternatively denoise temporal (column) sequences and spatial (row) sequences $N$ times. Each time, row or column sequences are denoised and inpainted (Fig.\ref{fig:denoising_inpainting}). By denoising along both spatial and temporal directions, we obtain an inpainted latent $\VidLatent_0$ which can be decoded into temporally smooth and semantically consistent sequences.}
    \label{fig:pipeline}
    \vspace{-0.1in}
\end{figure*}

However, this naive approach yields suboptimal results, as independently inpainting right-view frames without referring to left-view frames introduces semantic inconsistencies across views ({Fig.~\ref{fig:mutal_denoising} (a)}).
To address this, we propose a novel \textit{frame matrix} representation that contains frame sequences observed from multiple virtual viewpoints uniformly distributed along the interocular axis (Fig.~\ref{fig:viewpoints}, left). The frame matrix simultaneously encodes temporal and spatial information: each row corresponds to a video sequence with continuous camera motion at a fixed timestamp (spatial direction), while each column captures dynamic scene motions across time (temporal direction), as illustrated in Fig.~\ref{fig:pipeline} (second column). 
Capitalizing on the inherent video priors of scene and camera motions within the video generation model, we propose a joint optimization strategy that updates the entire \textit{frame matrix} along both temporal and spatial dimensions. During the denoising process, we employ resampling techniques~\cite{lugmayr2022repaint} to alternately refine frame sequences along the temporal and spatial directions, progressively enhancing both temporal stability and view coherence ({Fig.~\ref{fig:mutal_denoising} (c)}). The final 3D stereoscopic video is constructed by selecting the leftmost and rightmost frame sequences as the left-eye and right-eye view videos, respectively. 
Since high resolution is desirable in VR applications, we also explore stereoscopic video up-sampling in our method.

Beyond stereoscopic videos with fixed viewpoints, we further develop our method to create spatial videos enabling dynamic viewpoint changes. 
This is achieved by initializing the frame matrix's spatial dimension with multiple virtual viewpoints surrounding the input reference video (Fig.~\ref{fig:viewpoints}, right), followed by optimizing the inpainted frame matrix into a 4D Gaussian representation~\cite{yang2024deformable} to support downstream stereoscopic novel view synthesis. At each timestamp, the 3D geometry can be optimized using multi-view images with manually defined camera poses. Nevertheless, modeling continuous scene changes over time remains challenging due to the non-trivial task of estimating camera poses from a monocular video. To address this, we circumvent cross-time camera pose estimation by transferring camera motions into scene motions and modeling scene dynamics via learning time-dependent deformation offsets for each Gaussian entity. This representation naturally accommodates viewpoint changes while maintaining temporal continuity.

Furthermore, we observe that the inevitable resolution downsampling in latent-space video generation models is detrimental to the video inpainting task. During encoding, dark disoccluded pixels propagate through the network architecture, leading to corrupted boundary features and visually apparent artifacts, as demonstrated in Fig.~\ref{fig:update_feature}.
Contrary to conventional inpainting methods~\cite{avrahami2023blended} that perform a single encoding of masked images into latent space, our approach implements an iterative dual-space refinement and re-injection strategy. Throughout the denoising process, we simultaneously update both the disoccluded regions in image space and their corresponding latent feature representations. By re-encoding the progressively improved images, this approach purifies contaminated boundary features, leading to high-quality, artifact-free inpainting results.


To validate the efficacy of our method, we generate stereoscopic and spatial videos from monocular inputs produced by Sora~\cite{videoworldsimulators2024}, Lumiere~\cite{bar2024lumiere}, WALT~\cite{gupta2023photorealistic}, and Zeroscope~\cite{wang2023modelscope}. Comprehensive qualitative and quantitative evaluations demonstrate that our approach consistently outperforms existing baselines in immersive 3D video generation across multiple metrics. The key contributions of our work can be summarized as follows:
\begin{itemize}
    \item We design a novel stereoscopic and spatial video generation pipeline that eliminates the need for camera pose estimation or dataset-specific fine-tuning. To the best of our knowledge, our lines of work are pioneers in leveraging video diffusion models to facilitate the creation of immersive videos.
    \item We propose a novel \textit{frame matrix} representation that regularizes the diffusion-based video inpainting to generate semantically coherent and temporally consistent content.
    \item We meticulously develop a warping pipeline to provide correct occlusion relationships and propose a re-injection scheme that effectively mitigates the adverse effects of disoccluded regions in the latent space.
    \item We optimize the generated discrete observations into a continuous 4D Gaussian representation, enabling novel-time and novel-view stereoscopic video synthesis. By treating each Gaussian as a dynamic entity, we circumvent the challenging cross-time camera pose estimation.  
    \item We conduct comprehensive experiments that show the superiority of our approach over previous methods for immersive 3D video generation. 
\end{itemize}

\noindent\textbf{Difference to Our Conference Paper.}
This manuscripts substantially extend the ICLR 2025 conference paper~\cite{dai2024svg} in the following aspects: (1) \textit{Spatial video generation.} Unlike the conference version, which focuses on generating stereoscopic videos by taking the leftmost and rightmost columns of the denoised frame matrix, we optimize the full frame matrix, with cameras surrounding the input view, into a 4D Gaussian representation, which enables the generation of immersive spatial videos that support viewpoint changes. This optimization eliminates the need for cross-time camera pose estimation by modeling each Gaussian as a dynamic entity. (2) \textit{Partially observed objects.} Partially observed objects in monocular videos often lead to erroneous occlusion relationships when warping to a new viewpoint, resulting in incomplete content in the final result. To address this challenge, we explore the video outpainting, which provides more complete shapes, and propose an outpainting-then-inpainting design to produce results with complete objects. (3) \textit{Stereoscopic video super resolution.} 
We provide a stereoscopic video upsampling scheme that explicitly maintains the consistency between the left and right views during the video upsampling process.

\section{Related Work}
\label{sec:related_work}
\noindent\textbf{Video Generation.}
Video generation has witnessed remarkable advancements since the introduction of diffusion models~\cite{sohl2015deep, ho2020denoising}. Current state-of-the-art approaches~\cite{wang2023modelscope,videoworldsimulators2024,bar2024lumiere,gupta2023photorealistic,harvey2022flexible,ho2022imagen,ho2022video,singer2022make} typically address the challenge of limited annotated video data by extending pre-trained image generation models~\cite{rombach2022high,saharia2022photorealistic,ramesh2022hierarchical} through the insertion of temporal layers followed by video-domain fine-tuning~\cite{guo2023animatediff,blattmann2023align,wu2023tune}. Recent innovations in computational efficiency, exemplified by WALT~\cite{gupta2023photorealistic} and Lumiere~\cite{bar2024lumiere}, employ joint temporal-spatial compression to enable longer video generation. The field has further progressed with Sora~\cite{videoworldsimulators2024}, which leverages a transformer-based diffusion architecture~\cite{peebles2023scalable} trained on massive video datasets to achieve unprecedented generation quality. While existing work predominantly focuses on enhancing monocular video quality and duration, our research explores an orthogonal direction: harnessing these pre-trained video generation models for high-quality stereoscopic and spatial video synthesis. Recently, some works explored the use of stereo videos for mono-to-stereo conversion~\cite{zhao2024stereocrafter}, and object motion learning~\cite{jin2024stereo4d}; but the multi-view videos required for spatial video creation remain scarce.

\vspace{0.1in}
\noindent\textbf{Novel View Synthesis.}
Novel view synthesis has advanced significantly for both static and dynamic scenes~\cite{mildenhall2021nerf,yoon2020novel,li2022neural,kerbl20233d,muller2022instant,tucker2020single,han2022single,wang2023learning}. Early work~\cite{tucker2020single} demonstrated single-image view synthesis through multi-plane representations, while \cite{mildenhall2021nerf} introduced neural radiance fields (NeRF) that revolutionized static scene rendering through volumetric representations. Subsequent approaches extended these capabilities to dynamic scenes by incorporating deformation fields~\cite{park2021nerfies,huang2023sc,park2021hypernerf} or scene flow estimation~\cite{li2021neural}. Alternative paradigms emerged with methods like DynIBaR~\cite{li2023dynibar}, which leveraged motion fields and frame-based rendering, and RoDynRF~\cite{liu2023robust} that jointly optimized scene geometry and camera poses. The field further progressed with plane-based representations as shown in FVS~\cite{lee2023fast}, enabling efficient novel view video synthesis. While these methods support stereoscopic view synthesis and achieve impressive results, they remain fundamentally constrained by their dependence on accurate camera pose estimation and limited ability to synthesize unseen content. Our work breaks from this paradigm by eliminating the need for explicit pose estimation and introducing content hallucination capabilities, enabling stereoscopic and spatial video generation in challenging scenarios where traditional approaches fail.

\vspace{0.1in}
\noindent\textbf{Content Creation and Inpainting.}
Automated 3D content creation~\cite{hollein2023text2room,dai2024go,gao2024gaussianflow,yu2023wonderjourney,chen2024v3d} represents another relevant research direction. Current approaches employ various strategies, including inpainting and multi-view generation~\cite{liu2023zero,wang2024stereodiffusion,zuo2024videomv}. In particular, some works converted RGB-D images into layered depth images for 3D photo creation~\cite{shih20203d, jampani2021slide}, enabling observations from different viewpoints. To enlarge the viewpoint changes, Text2Room~\cite{hollein2023text2room} demonstrated room-scale 3D generation by progressively warping images into novel views and using text-guided inpainting for disocclusions, while WonderJourney~\cite{yu2023wonderjourney} automated this process through large language model integration. Although pretrained video inpainters~\cite{zhou2023propainter,li2022towards} could theoretically extend these approaches to dynamic content, they struggle to produce high-quality, consistent 3D results. Deep3D~\cite{xie2016deep3d} offers stereoscopic video conversion but requires proprietary 3D movie data and lacks flexibility for creative applications like adjustable stereo baselines. Our work explores an alternative paradigm: leveraging video generation models for immersive 3D video creation without requiring specialized training datasets.

\section{Method}
\subsection{Overview}
\noindent\textbf{Stereoscopic Video Generation.} 
Conditioned on a text prompt or a single image $\Prompt$, our method aims to generate immersive 3D video $\{\Vid_{\LeftSub},\Vid_{\RightSub}\}$, consisting of two monocular sequences representing left and right views. 
The generated videos should satisfy the following three critical properties: (1) the appearance and semantics between the left $\Vid_{\LeftSub}$ and right $\Vid_{\RightSub}$ views should be consistent and be temporally stable, (2) prominent and immersive 3D effects, and (3) diverse yet controllable output conditioned on the input.

In this work, we propose a training-free pipeline that leverages an off-the-shelf depth estimator~\cite{yang2024depth} and a pretrained monocular video diffusion model $\Generator$ (e.g., Zeroscope~\cite{wang2023modelscope}) for 3D stereoscopic video generation. Our pipeline begins by generating a monocular video for the left view using a video diffusion model (Eq.~\ref{eq:leftgen}, $\Noise_t \sim \mathcal{N}(\mathbf{0}, \mathbf{I})$ denotes the Gaussian noise sampled at diffusion timestep $t$). To produce the corresponding right view while maintaining 3D consistency, we first estimate depth $\VidDepth_\LeftSub$ from the left view video, then apply stereoscopic warping~\cite{wang2023learning,han2022single} to obtain the initial right view sequence $\Vid_{\LeftSub\rightarrow\RightSub}$ and its disocclusion masks $\ImageMasks_\RightSub$ (Eq.~\ref{eq:warp}). The disoccluded regions are then completed through a diffusion-based inpainting process (Eq. \ref{eq:inpaint})~\cite{avrahami2023blended,lugmayr2022repaint}, where $\Generator$ synthesizes plausible content to generate the final right-view video $\Vid_{\RightSub}$. 

\vspace{-0.1in}
{\small
\begin{align}
    &\Vid_{\LeftSub} = \Generator(\{\Noise_t | t=1,..., T\}, c), \label{eq:leftgen} \\
    &\Vid_{\LeftSub\rightarrow\RightSub}, \ImageMasks_\RightSub = \textrm{Warp}_{\LeftSub\rightarrow\RightSub}(\Vid_{\LeftSub}, \VidDepth_\LeftSub),
    \label{eq:warp} \\
    &\Vid_{\RightSub} = \Generator(\{\Noise_t | t=1,..., T\}, c, \Vid_{\LeftSub\rightarrow\RightSub}, \ImageMasks_\RightSub).
    \label{eq:inpaint}
\end{align}}

\begin{figure}
    \centering
    \includegraphics[width=1.0\linewidth]{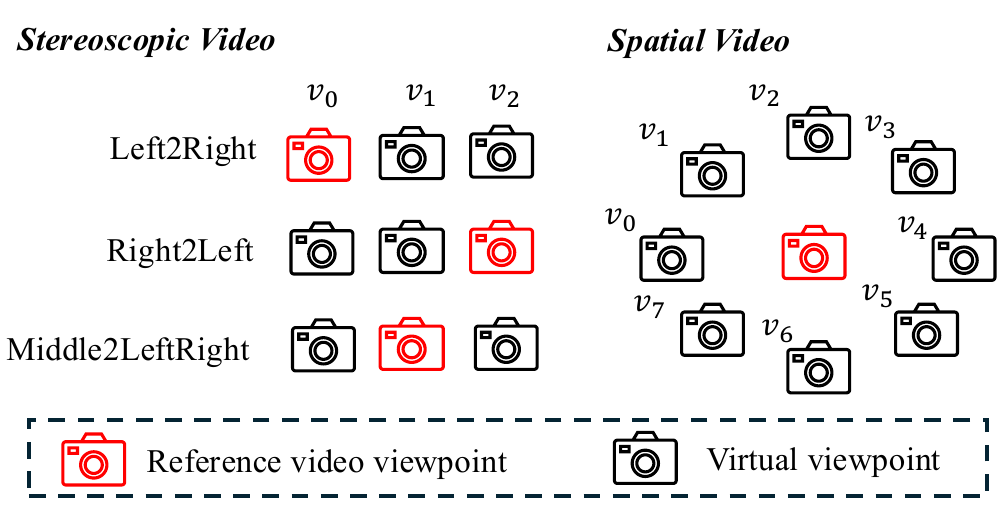}
    \vspace{-0.25in}
    \caption{\textbf{The setting of virtual camera viewpoints.} For stereoscopic video generation, virtual cameras are distributed along the interocular baseline. We use the left2right scheme by default, but other schemes are also viable. For spatial video generation, we position multiple virtual cameras around the reference viewpoint of the monocular video. To ensure circular consistency, the first and last columns of the \textit{frame matrix} share the same first camera viewpoint.}
    \label{fig:viewpoints}
    \vspace{-0.1in}
\end{figure}

\noindent\textbf{Spatial Video Generation.} 
Apple has introduced the concept of spatial video. Unlike conventional stereoscopic videos that provide fixed viewpoints, spatial video enables dynamic viewpoint changes, offering a more immersive and interactive viewing experience. To address disocclusions arising from viewpoint changes, we propose warping the monocular input video $\Vid_{l}$ into multiple camera viewpoints (Eq.~\ref{eq:warp_m}) surrounding the input monocular video (Fig.~\ref{fig:viewpoints}, right) and employing a video diffusion model to hallucinate these disoccluded regions (Eq.~\ref{eq:inpaint_m}). These videos maximally cover potential unknown areas, which are then optimized into a 4D representation for efficient stereoscopic novel view rendering (Eq.~\ref{eq:gaussian_m}).

\vspace{-0.1in}
{\small
\begin{align}
    &\Vid_{warp}, \ImageMasks = \textrm{Warp}_{\LeftSub\rightarrow\textrm{m}}(\Vid_{\LeftSub}, \VidDepth_\LeftSub),
    \label{eq:warp_m} \\
    &\Vid = \Generator(\{\Noise_t | t=1,..., T\}, c, \Vid_{\textrm{warp}}, \ImageMasks),
    \label{eq:inpaint_m} \\
    &(\Vid_{l\_spatial}, \Vid_{r\_ spatial}) := \mathbb{G}(\Vid).
    \label{eq:gaussian_m}
\end{align}}

Our framework comprises five technical components: (1) depth-based video warping (Sec.~\ref{sec:video_warp}), (2) a novel \textit{frame matrix} representation for video inpainting (Sec.~\ref{sec:frame_matrix}), where the frame matrix representation significantly enhances both the semantic coherence across viewpoints and the temporal consistency across frames, (3) a \dualupdate~mechanism for boundary refinement (Sec.~\ref{sec:boundary_reinjection}), (4) a pipeline to extract stereoscopic video and spatial video from inpainted frame matrix (Sec.~\ref{sec:4d_video}), and (5) an optional stereoscopic video super-resolution scheme (Sec.~\ref{sec:video_sr}).
Fig.~\ref{fig:pipeline} provides an overview of our pipeline.

\subsection{Depth-Based Monocular Video Warping}
\label{sec:video_warp}
We employ the depth estimation model of~\cite{yang2024depth} to predict per-frame depth values, which are subsequently temporally smoothed to enhance consistency across the video frames. Specifically, consecutive depth frames are aligned using estimated optical flows from~\cite{teed2020raft}, while outliers in the predicted depths are suppressed via temporal convolution with a Gaussian kernel. The resulting RGB-D frames are then warped into pre-defined camera viewpoints. However, warped images often exhibit isolated pixels and entangled background and foreground (Fig.~\ref{fig:ab_data_preprocessing}, left), which compromises visual quality~\cite{dai2020neural}. To address these artifacts, we propose to warp RGB-D images onto multi-plane images~\cite{zhou2018stereo}, and then eliminate isolated pixels and cracks to obtain clean warped images for the subsequent generative inpainting. 

\textit{Multi-Plane Projection.} 
Given the RGB-D images, we warp them into a target camera view. Rather than projecting all pixels onto a single image plane and resolving occlusions via z-buffering, we partition the camera view space into $N$ discrete planes $\{\Vid_{warp\_1}^{\text{s1}}, \dots, \Vid_{warp\_K}^{\text{s1}}\}$ (where $K=4$ in our implementation), stratified according to near and far depth bounds. Each pixel is projected onto its closest image plane based on depth values. Binary masks $\{\ImageMasks_1^{\text{s1}}, \dots, \ImageMasks_K^{\text{s1}}\}$ are used to identify occupied pixel positions on each plane. This multi-plane decomposition naturally separates foreground and background elements into distinct layers. Consequently, common warping artifacts, including isolated pixels and foreground-background entanglement, become significantly more tractable to address.

\textit{Remove Isolated Points.} 
Due to depth inaccuracies near image content boundaries, boundary pixels are often warped to incorrect positions, resulting in isolated pixels (see red box in Fig.~\ref{fig:ab_data_preprocessing}, left). We leverage the observation that isolated pixels typically have few or no neighboring pixels. For detection, we convolve each mask image $\ImageMasks_i^{\text{s1}}$ with a $3\times3$ kernel and empirically classify pixels as isolated when their convolved values fall below 0.5. These isolated pixels are subsequently removed from both RGB images and masks, yielding the updated sets $\{\Vid_{warp\_1}^{\text{s2}}, \dots, \Vid_{warp\_K}^{\text{s2}}\}$ and $\{\ImageMasks_1^{\text{s2}}, \dots, \ImageMasks_K^{\text{s2}}\}$.

\textit{Fill Cracks.} The non-watertight nature of depth images often leads to cracks and holes in warped results, causing foreground-background confusion (e.g., visible road texture through the dog's ear in Fig.~\ref{fig:ab_data_preprocessing}, left). Building upon our isolated pixel removal approach, we detect cracks by convolving each mask $\ImageMasks_i^{\text{s2}}$ with a $3\times3$ Gaussian kernel. In this paper, positions with no pixel values (0 in $\ImageMasks_i^{s2}$) but with values greater than 0.2 after convolution are empirically considered as cracks. The detected cracks are filled by local pixel interpolation, producing the final multi-plane images $\{\Vid_{warp\_1}^{\text{s3}}, \dots, \Vid_{warp\_K}^{\text{s3}}\}$ and the corresponding masks $\{\ImageMasks_1^{\text{s3}}, \dots, \ImageMasks_K^{\text{s3}}\}$.

After processing all individual planes for artifacts, we composite them into the final output $\Vid_{warp}$ through back-to-front blending. The blending operation follows:
\begin{equation}
    \small
    \Vid_{warp} = \Vid_{warp}\times(1-\ImageMasks_i^{s3}) + \Vid_{warp\_i}^{s3}\times \ImageMasks_i^{s3},\ for\ i\ in\ [K, ..., 1],
    \label{eq:blend}
\end{equation}
where front plane content ($i$ close to 1) naturally occludes back plane content ($i$ close to $K$). These warped images $\Vid_{warp}$ will be processed via video inpainting.

Nevertheless, monocular videos often contain partially visible objects, as shown by the coconut tree in Fig.~\ref{fig:incomplete} (a). When depth-based warping is directly applied to such cases, it results in incomplete objects in the warped images that cannot be adequately repaired through video inpainting. To resolve this issue, we employ video outpainting (implemented via single video denoising inpainting as described in Sec.~\ref{sec:frame_matrix}) to extend the visible content, thereby providing more complete objects for subsequent warping. The outpainting padding size $P$ is set equal to the maximum disparity value, computed as:
\begin{equation}
P = \frac{f \times b}{d_{\text{min}}},
\end{equation}
where $f$ is the focal length, $b$ represents the binocular baseline, and $d_{\text{min}}$ is the minimum scene depth.

\subsection{Video Inpainting with Frame Matrix}
\label{sec:frame_matrix}
The inpainting pipeline serves as a crucial component for maintaining both spatial/semantic coherence and temporal consistency in the output video. Although existing image inpainting methods~\cite{avrahami2023blended,lugmayr2022repaint} offer a viable starting point, they inherently fail to preserve stability across both spatial and temporal dimensions. Therefore, we introduce a \textit{frame matrix} representation, which addresses both issues. In the following, we first introduce the basic idea of single video denoising inpainting, followed by advanced frame matrix denoising inpainting. 

\begin{figure*}
    \centering
    \includegraphics[width=1.0\linewidth]{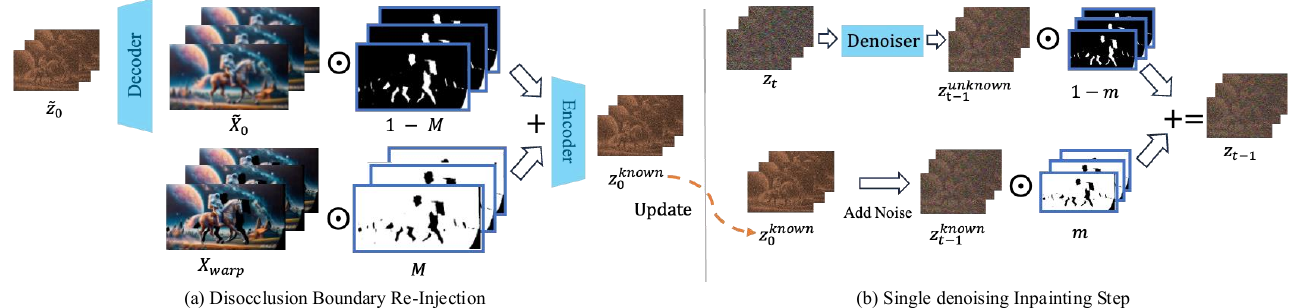}
    \vspace{-0.2in}
    \caption{\textbf{Denoising inpainting}. This figure explains the operations in the purple box of Fig.\ref{fig:pipeline}. (a) We re-inject the generated content from a denoised latent $\widetilde{\VidLatent}_0$ to update $\VidLatent_0^{known}$, reducing its feature corruption on the disocclusion boundary. (b) A noisy latent $\VidLatent_t$ is denoised to $\VidLatent_{t-1}^{unknown}$. We combine its disoccluded region with the noisy known region of $\VidLatent_{t-1}^{known}$ to obtain a complete denoised latent feature $\VidLatent_{t-1}$. }
    \label{fig:denoising_inpainting}
\end{figure*}

\vspace{0.1in}
\noindent\textbf{Single Video Denoising Inpainting.} 
Inspired by RePaint~\cite{lugmayr2022repaint}, we extend the diffusion-based image inpainting to video inpainting. We use the video generation model $\Generator$ (i.e., Zeroscope~\cite{wang2023modelscope}) as our inpainting tool, which is a latent diffusion model consisting of a VAE encoder $\Encoder$, a decoder $\Decoder$ and a latent denoiser $\{ \epsilon_{\theta}, \Sigma_{\theta}\}$. First, the warped video is fed into the VAE encoder to obtain video latent features $\VidLatent_0^{\text{known}} = \Encoder(\Vid_{\LeftSub\rightarrow\RightSub})$. Accordingly, the image disocclusion masks $\ImageMasks_\RightSub$ are resized to match the resolution of encoded latent features, yielding latent disocclusion masks $\LatentMasks$. The denoising process initializes with a random noisy latent map $\VidLatent_T \sim~ \mathcal{N}(\mathbf{0}, \mathbf{I})$. At each denoising step $t$, we denoise the latent map $\VidLatent_t$ according to Eq.~\ref{eq:inpaintunknown}, sample a subsequent noisy latent map from $\VidLatent_0$ following Eq.~\ref{eq:inpaintknown} and combine them with $\LatentMasks$ to obtain the $\VidLatent_{t-1}$ using Eq.~\ref{eq:inpaintt_1}. We visualize the above steps in Fig.\ref{fig:denoising_inpainting} (b):

\vspace{-0.1in}
{\small
\begin{align}
    \VidLatent_{t-1}^{\text{known}} \sim&~ \mathcal{N}\left(\sqrt{\NoiseScale}\VidLatent_0^{\text{known}}, (1 - \NoiseScale)\mathbf{I}\right),
    \label{eq:inpaintknown} \\
     \VidLatent_{t-1}^{\text{unknown}} \sim&~ \mathcal{N}\left(\frac{1}{\sqrt{1-\beta_t}}\left(\VidLatent_t - \frac{\beta_t}{\sqrt{1-\Bar{\alpha}_t}}\epsilon_{\theta}(\VidLatent_t,\Prompt,t)\right), \Sigma_{\theta}(\VidLatent_t, \Prompt, t)\right),
    \label{eq:inpaintunknown} \\
    \VidLatent_{t-1} =& ~\LatentMask ~\odot~ \VidLatent_{t-1}^{\text{known}} + (1-\LatentMask) ~\odot~ \VidLatent_{t-1}^{\text{unknown}},
    \label{eq:inpaintt_1}
\end{align}}
where $\NoiseScale$ represents the total noise variance and $\beta_t$ denotes the one-step noise variance at timestep $t$. The denoising process employs two learned functions: $\epsilon_{\theta}(\VidLatent_t,\Prompt,t)$ predicts the noise component, while $\Sigma_{\theta}(\VidLatent_t,\Prompt,t)$ estimates the variance for the latent map at timestep $t-1$. The final inpainted sequence $\Frame_\RightSub$ is obtained by decoding the denoised latent representation through $\Decoder(\VidLatent_0)$. Considering that the VAE decoder cannot perfectly reconstruct warped pixels, we adopt Poisson Blending~\cite{perez2023poisson} to enhance the transition between generated and warped content in image space.

By applying the above video inpainting design to the warped images, we successfully hallucinate occluded regions while preserving known image content. Although the video diffusion model ensures temporal smoothness, the inpainted content on the warped view usually lacks semantic coherency with respect to the given reference view, as shown in the third column of Fig.~\ref{fig:sematic_matching}. This is because we abandon the conditioning on the reference view and process the warped view independently

\vspace{0.1in}
\noindent\textbf{Frame Matrix Denoising Inpainting.} 
We introduce a novel representation—the \textit{frame matrix}—designed to ensure consistent dynamic content generation across both space and time. As illustrated in Fig.~\ref{fig:pipeline} (top), the frame matrix is structured as a two-dimensional array of frames, where each row corresponds to multiple camera perspectives captured at the same timestamp, while each column represents a temporal sequence of frames recorded from a fixed camera viewpoint. Formally, the frame matrix can be defined as:
\[\tiny{
\Vid \equiv 
\left[
  \begin{array}{ccc}
    \vertbar &         & \vertbar \\
    \Vid_{(:,0)}    &  \ldots & \Vid_{(:,\ViewSubMax)}    \\
    \vertbar &         & \vertbar 
  \end{array}
\right]
 \equiv
\left[
  \begin{array}{ccc}
    \horzbar & {\Vid_{(0,:)}} & \horzbar \\
             & \vdots    &          \\
    \horzbar & \Vid_{(\FrameSubMax,:)} & \horzbar
  \end{array}
\right],}
\]
where $\FrameSubMax$ and $\ViewSubMax$ denote the largest indices of time stamps and camera views, respectively. A spatial sequence (row) $\Vid_{(\FrameSub,:)}$ forms a video with camera motions, while a temporal sequence (column) $\Vid_{(:,\ViewSub)}$ forms a video with time-varying scene motions. Since the video generation model can generate temporally and semantically consistent dynamic or static videos, we adopt it to jointly denoise the rows and columns of \textit{frame matrix}, ensuring spatial and temporal consistency. Finally, we can extract the immersive 3D video from it.

To initialize the frame matrix, we place $\ViewSubMax$ camera views with the same orientation of the reference view along the interocular baseline for stereoscopic setups and distribute camera views surrounding the reference view for spatial setups (Fig.~\ref{fig:viewpoints}). Next, the depth-based warping method (Sec.~\ref{sec:video_warp}) is employed to warp the reference video into these manually defined viewpoints and obtain $\Vid_{warp} \equiv [\Vid_{(:,0)}, \Vid_{(:,1)}, ..., \Vid_{(:,\ViewSubMax)}]$ with disocclusion masks $\ImageMasks$. The definition of the camera trajectory is flexible, depending on the demands.

\begin{figure}[!htb]
    \centering
    \includegraphics[width=1.0\linewidth]{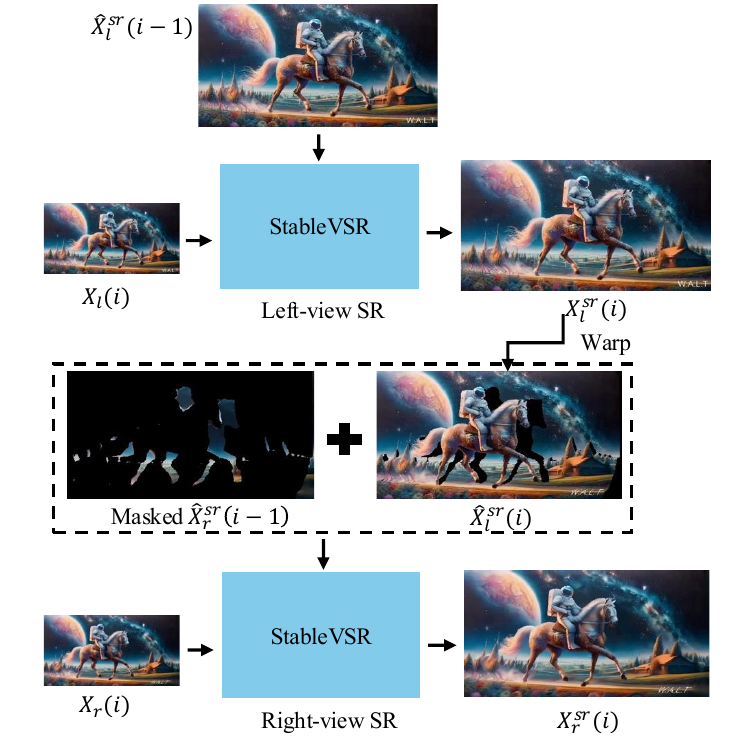}
    \vspace{-0.2in}
    \caption{\textbf{Stereoscopic video super resolution.} The left-view super resolution is conditioned on the previous upsampled frame, while the right-view super resolution is conditioned on both left-view and previous upsampled frames.}
    \label{fig:sr_pipeline}
    \vspace{-0.1in}
\end{figure}

\begin{figure*}[!htb]
    \centering
    \includegraphics[width=1.0\linewidth]{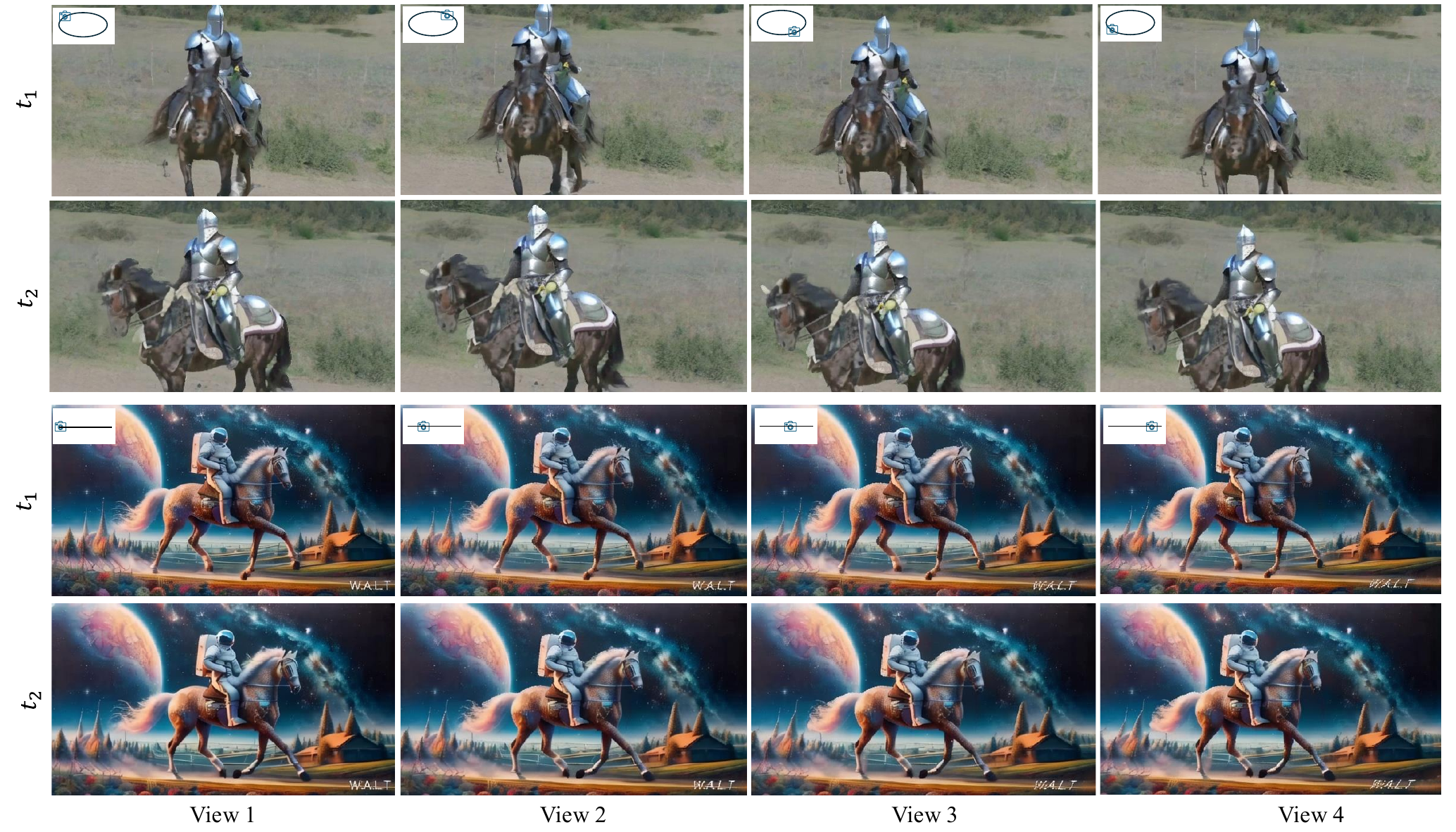}
    \vspace{-6mm}
    \caption{\textbf{Frames sampled from denoising inpainted frame matrix.} Each row represents a video with camera motions at a fixed timestamp. Each column is a video, recording the scene motion from a camera viewpoint.}
    \label{fig:multi_view_traj}
    \vspace{-0.1in}
\end{figure*}

The denoising process of frame matrix is similar to single video inpainting, we first encode the warped frame matrix into latent space as $\VidLatent_0^{\text{known}} = \Encoder(\Vid_{warp})$, while resizing the disocclusion masks $\ImageMasks$ to obtain the latent mask matrix $\LatentMasks$. We initialize the latent noise as $\VidLatent_T \sim~ \mathcal{N}(\mathbf{0}, \mathbf{I})$. 
As illustrated in Fig.~\ref{fig:pipeline} (Bottom), at each denoising timestep $t$, we extend the resampling mechanism~\cite{lugmayr2022repaint} to alternatively denoise column sequences and row sequences $N$ times. Each time, row or column sequences are denoised following Eq.~\ref{eq:inpaintknown}-\ref{eq:inpaintt_1} and then re-add noise back to the previous denoising timestep: 

\vspace{-0.1in}
{\small
\begin{align}
    \VidLatent_{t} \sim \mathcal{N}(\sqrt{1-\beta_{t-1}}\VidLatent_{t-1}, \beta_{t-1}\mathbf{I}).
    \label{eq:add_resampling_noise}
\end{align}}

\begin{algorithm}[!htb]
  \caption{Frame Matrix Denoising Inpainting}
  \label{algo_fm}
  \begin{algorithmic}
    \STATE \textbf{Input: } $\VidLatent_T \sim~ \mathcal{N}(\mathbf{0}, \mathbf{I})$: Initial noisy latent maps \\
    $\VidLatent_0$: Initial clean latent maps \\
    \FOR{$t=T,...,1$}
        \FOR{$n=1,...,N$}
            \IF{n is odd}
                \STATE Denoise time sequences
                $\{\VidLatent_{(:,\ViewSub)t} | \ViewSub = 0,..., \ViewSubMax\}$:
                \FOR{$\ViewSub = 0,..,\ViewSubMax$}
                   \STATE $\VidLatent_{(:,\ViewSub)t-1}^{\text{known}} \sim \mathcal{N}(\sqrt{\NoiseScale}\VidLatent_{(:,\ViewSub)0}, (1 - \NoiseScale)\mathbf{I})$ \\
                   \STATE $\VidLatent_{(:,\ViewSub)t-1}^{\text{unknown}} \sim \mathcal{N}(\mu_{\theta}(\VidLatent_{(:,\ViewSub)t}, \Prompt, t), \Sigma_{\theta}(\VidLatent_{(:,\ViewSub)t}, \Prompt, t))$ \\
                    \STATE $\VidLatent_{(:,\ViewSub)t-1} = \LatentMasks_{(:,\ViewSub)} ~\odot~ \VidLatent_{(:,\ViewSub)t-1}^{\text{known}} + (1-\LatentMask_{(:,\ViewSub)}) ~\odot~ \VidLatent_{(:,\ViewSub)t-1}^{\text{unknown}}$
                \ENDFOR
            \ELSE
            \STATE Denoise view sequences $\{\VidLatent_{(\FrameSub,:)t} | \FrameSub = 0,..., \FrameSubMax \}$:
                \FOR{$\FrameSub = 0,..,\FrameSubMax$}
                   \STATE $\VidLatent_{(\FrameSub,:)t-1}^{\text{known}} \sim \mathcal{N}(\sqrt{\NoiseScale}\VidLatent_{(\FrameSub,:)0}, (1 - \NoiseScale)\mathbf{I})$ \\
                   \STATE $\VidLatent_{(\FrameSub,:)t-1}^{\text{unknown}} \sim \mathcal{N}(\mu_{\theta}(\VidLatent_{(\FrameSub,:)t}, \Prompt, t), \Sigma_{\theta}(\VidLatent_{(\FrameSub,:)t}, \Prompt, t))$ \\
                    \STATE $\VidLatent_{(\FrameSub,:)t-1} = \LatentMasks_{(\FrameSub,:)} ~\odot~ \VidLatent_{(\FrameSub,:)t-1}^{\text{known}} + (1-\LatentMask_{(\FrameSub,:)}) ~\odot~ \VidLatent_{(\FrameSub,:)t-1}^{\text{unknown}}$
                \ENDFOR
            \ENDIF

            \IF{$n<N$}
             \STATE Add back one noise step for resampling:\\
                        \STATE $\VidLatent_{t} \sim \mathcal{N}(\sqrt{1-\beta_{t-1}}\VidLatent_{t-1}, \beta_{t-1}\mathbf{I})$
            \ENDIF
            
        \ENDFOR
    \ENDFOR
  \end{algorithmic}
\end{algorithm}
Algorithm~\ref{algo_fm} presents detailed steps for denoising the \textit{frame matrix} via spatial-temporal resampling. {{By denoising along these two directions alternatively, temporal and spatial priors are balanced against each other to predict results with enhanced consistency (Fig.~\ref{fig:mutal_denoising})}}.

\vspace{0.1in}
\noindent\textbf{More Analysis of Frame Matrix.}
\textit{(1) High-level perspective.} In practical 3D stereoscopic video production, we typically record with two cameras to obtain the video sequences. Since both cameras capture the same scene, this means that gradually moving the position of the left camera towards the position of the right camera can also produce a coherent spatial-direction video. This principle applies equally to 3D stereoscopic video generation: we must jointly consider both the temporal and spatial directions to ensure the generated perspectives properly represent the same scene. Furthermore, our experiments show that gradually expanding the inpainting region along the spatial direction yields more stable and reasonable results than attempting to inpaint large missing regions in a single step. This progressive approach better maintains consistency across viewpoints.
\textit{(2) Mathematical perspective.} Each frame $\VidLatent_{t-1}(i, j)$ in the latent \textit{frame matrix} should be as consistent as the denoising results from both temporal and spatial directions. Mathematically:

\vspace{-0.1in}
{\small
\begin{align}
    L(i,j) = ||\VidLatent_{t-1}(i,j) - \VidLatent_{t-1}^{\mathbb{T}}(i,j)||_2 +& ||\VidLatent_{t-1(i,j)} - \VidLatent_{t-1}^{\mathbb{S}}(i,j)||_2, 
    \label{eq:alterd_part1} \\
    \VidLatent_{t-1}^{\mathbb{T}}=\Theta^{\mathbb{T}}(\VidLatent_t), \quad\VidLatent_{t-1}^{\mathbb{S}}&=\Theta^{\mathbb{S}}(\VidLatent_t),    
    \label{eq:alterd_part2}
\end{align}}
where $\Theta$ is a pre-trained video diffusion model, $\Theta^{\mathbb{S}}$ and $\Theta^{\mathbb{T}}$ indicate denoising operations along the spatial and temporal dimensions, respectively. Here, $L$ is a quadratic Least-Squares (LS) problem where the solution closely approximates all diffusion samples $z^{\mathbb{S}}_{t-1}(i,j)$ and $z^{\mathbb{T}}_{t-1}(i,j)$. In practice, we perform denoising in both spatial and temporal directions to approximate optimal results. 

\subsection{\DualUpdate}
\label{sec:boundary_reinjection}
Since most video generation models are using latent diffusion, the disoccluded regions of $\Vid_{warp}$ will be propagated beyond the latent mask $\LatentMasks$ during VAE encoding (\textit{e.g.}, Zeroscope downsamples by $8\times$), leading to corrupted latent features around $\VidLatent_0^{\text{known}}$'s disocclusion boundary. This would result in artifacts in the final results (Fig.~\ref{fig:update_feature}, left). 

To address this issue, we propose a latent feature refinement strategy that re-injects denoised information into the disocclusion boundary features. Specifically, we predict denoised latent features~\cite{ho2020denoising} and decode them into a denoised video (Eq.~\ref{eq:denoised0}). Next, we replace its unoccluded regions with warped pixels, producing a video that maintains faithfulness to the reference view while improving the disoccluded areas. Finally, re-encoding this composite video yields an updated latent representation $\VidLatent_0^{\text{known}}$ (Eq.~\ref{eq:update_latent}), effectively mitigating boundary corruption:

\vspace{-0.1in}
{\small
\begin{align}
    \widetilde{\Vid}_0 &= \Decoder(\widetilde{\VidLatent}_0), \text{where}~ \widetilde{\VidLatent}_0 = \frac{1}{\sqrt{\Bar{\alpha}_t}}\left(\VidLatent_t - \sqrt{1-\Bar{\alpha}_t}\epsilon_{\theta}(\VidLatent_t, \Prompt,t)\right)
    \label{eq:denoised0}, \\
    \VidLatent_{0}^{\text{known}} &= \Encoder\left(\ImageMasks ~\odot~ \Vid_{warp} + (1-\ImageMasks) ~\odot~ \widetilde{\Vid}_0\right).
    \label{eq:update_latent}
\end{align}}
This improved $\VidLatent_0^{\text{known}}$ will be used in Eq. \ref{eq:inpaintknown} for the next iteration.

\subsection{Extract Stereoscopic and Spatial Videos}
\label{sec:4d_video}
After denoising inpainting the latent frame matrix, we decode it into multi-view videos $\Vid = \Decoder(\VidLatent_0)$, where the stereoscopic and spatial videos can be extracted.  

\vspace{0.1in}
\noindent\textbf{Extract Stereoscopic Video.} Since stereoscopic videos provide a fixed viewpoint when observing a scene, we select the leftmost and rightmost columns from the frame matrix ($\Vid_l, \Vid_r$) to represent observations of the left and right eyes.

\vspace{0.1in}
\noindent\textbf{Extract Spatial Video.} 
{{The output frame matrix, composed of multi-view videos, is a discrete observation of the scene and cannot support the continuous synthesis (view or time) required by spatial videos. In addition, some jitters are retained in the generated frame matrix. To address these problems, we optimize the generated multi-view observations into a 4D representation, which serves as a global constraint, enabling continuous and consistent stereoscopic view synthesis. Considering rendering quality and efficiency, we choose \textit{Deformable Gaussian Splatting (DGS)}~\cite{yang2024deformable} as our backbone, which takes posed images as inputs and learns a set of 3D Gaussians $G(\mathbf{p}, \mathbf{r}, \mathbf{s}, \mathbf{\sigma}, \mathbf{c})$ in a canonical space, associated with time-dependent offsets $(\mathbf{\delta^t_p}, \mathbf{\delta_r^t}, \mathbf{\delta_s^t})$ to indicate Gaussian primitive's movements.}} Here, $\mathbf{p}$ denotes the center positions, $\mathbf{\sigma}$ indicates opacity, $\mathbf{c}$ represents view-dependent appearances, while the quaternion $\mathbf{r}$ and the scaling $\mathbf{s}$ define the shape of Gaussian primitives.

Specifically, we back-project the first RGB-D frame ($t=0$) of the reference video into the 3D canonical space, which serves as the initial state for the 3D Gaussian primitives. To model dynamic motions, the 3D Gaussian primitives are deformed $G(\mathbf{p+\delta_p^t}, \mathbf{r+\delta_r^t}, \mathbf{s+\delta_s^t}, \mathbf{\sigma}, \mathbf{c})$ based on time-dependent offsets, which are predicted using an $\textit{MLP}$ with given timestamp $t$ and positions $\mathbf{p}$ of 3D Gaussians as inputs:
\begin{equation}
    \small
     \left(\mathbf{\delta^t_p}, \mathbf{\delta_r^t}, \mathbf{\delta_s^t}) = \mathit{MLP}(\gamma(sg(\mathbf{p})),\gamma{(t)}\right),
    \label{eq:dff}
\end{equation}
where $sg$ indicates a stop-gradient operation, $\gamma$ denotes the positional encoding. {During the optimization process, the deformed 3D Gaussians at timestamp $t$ are rendered~\cite{kerbl20233d} into 2D images using pre-defined camera viewpoints (row t of \textit{frame matrix}) and compared with generated images $\Vid[t,:]$ at timestamp $t$, where their discrepancy $\mathcal{L}$ will guide the optimization of Gaussian parameters and the \textit{MLP} network. To ensure high-quality renderings, we use a combination of $\mathcal{L}_1$ loss and $LPIPS$ loss:}
\begin{equation}
    \small
     \mathcal{L} = \lambda_1\times \mathcal{L}_1 + \lambda_2\times LPIPS.
    \label{eq:loss}
\end{equation}
To improve robustness, we further augment this optimization process by randomly sampling viewpoints within the circular trajectory (Fig.~\ref{fig:viewpoints}) for supervision, where rendered unknown regions are excluded from loss computation. Following~\cite{kerbl20233d}, adaptive density control is employed, but we disable the pruning operation in practice to obtain renderings with higher quality.

Unlike the vanilla \textit{DGS}, our optimization process circumvents the need for challenging cross-time camera pose estimation by assuming camera poses across timestamps are fixed and modeling each Gaussian as a dynamic entity. In this way, temporal motion can be established by predicting inter-frame offsets, while multi-view images provide constraints on the 3D geometry of each timestamp.

\begin{figure*}[!htb]
    \centering
    \includegraphics[width=1.0\linewidth]{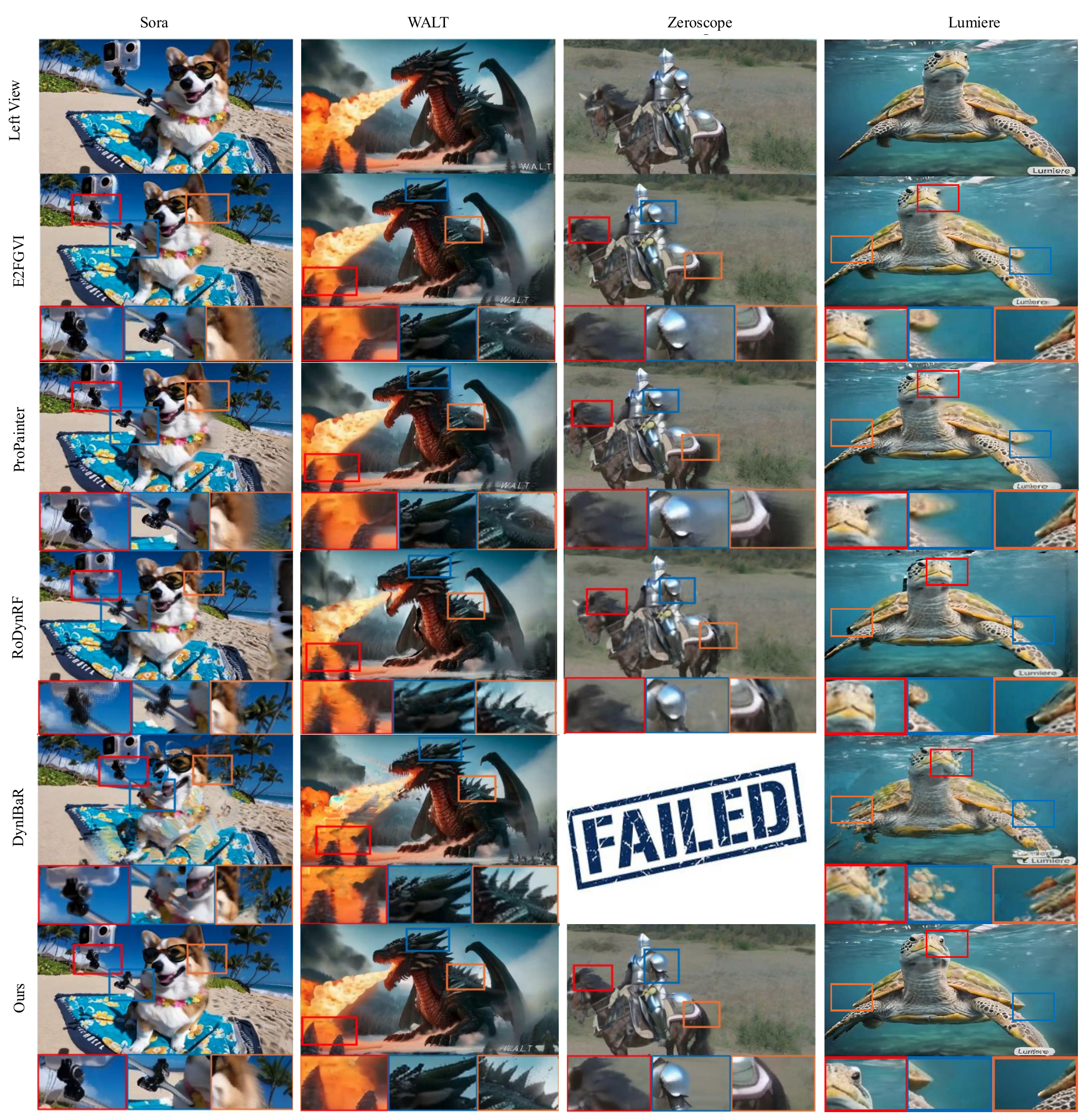}
    \vspace{-0.2in}
    \caption{\textbf{Qualitative comparisons on stereoscopic generation.} The first row is left-view images, while the others show the synthesized right-view images. The video inpainting methods E2FGVI and ProPainter tend to generate blurry content in disoccluded regions. RoDynRF and DynIBaR cannot generate occluded content and produce results with artifacts.}
    \label{fig:comparisons}
\end{figure*}

\subsection{Stereoscopic Video Super Resolution}
\label{sec:video_sr}
We develop a stereo video super-resolution framework built upon a pre-trained monocular video super-resolution model~\cite{rota2024enhancing} (denoted as $\mathcal{SR}$). As illustrated in Fig.~\ref{fig:sr_pipeline}, our pipeline first processes the left-view video sequence $\Vid_l(i)$ through the original $\mathcal{SR}$ architecture:
\begin{equation}
\small
\Vid_{l}^{sr}(i) = \mathcal{SR}\left(\Vid_l(i), \widehat{\Vid}_{l}^{sr}(i-1)\right),
\label{eq:sr_left}
\end{equation}
where $\widehat{\Vid}{l}^{sr}(i-1)$ represents the temporally aligned version (via optical flow) of the previous upsampled frame, which serves as a temporal condition to ensure inter-frame consistency. In contrast, the right-view upsampling incorporates both temporal condition from $\widehat{\Vid}{r}^{sr}(i-1)$ and cross-view condition from $\widehat{\Vid}_{l}^{sr}(i)$ to maintain both temporal stability and stereo coherence:
\begin{equation}
    \small
    \Vid_{r}^{sr}(i)  = \mathcal{SR}\left(\Vid_{r}(i), Mix\left(\widehat{\Vid}_{l}^{sr}(i),\widehat{\Vid}_{r}^{sr}(i-1)\right)\right),
    \label{eq:sr_right}
\end{equation}
where $Mix$ denotes an operation that replaces regions in the upsampled right-view frame $\widehat{\Vid}_{r}^{sr}(i-1)$ with their warped counterparts from the upsampled left-view frame $\widehat{\Vid}_{l}^{sr}(i)$.

\section{Experiments}
\label{sec:exp}

\subsection{Implementation Details}
\noindent\textbf{Datasets.}
To validate our method's effectiveness, we conduct experiments with a variety of recent video generation models, including Sora~\cite{videoworldsimulators2024}, Lumiere~\cite{bar2024lumiere}, WALT~\cite{gupta2023photorealistic}, and Zeroscope~\cite{wang2023modelscope}. These models generate diverse videos from a wide range of input text prompts, encompassing a broad spectrum of subjects, including humans, animals, buildings, and imaginary content.

\vspace{0.1in}
\noindent\textbf{Generation and optimization Details.}  \textit{1) Stereoscopic video.}
To achieve realistic 3D effects, we first normalize the predicted depth values from~\cite{yang2024depth} to the range $(1, 10)$ and set the interocular baseline to $0.07$ meter. Our multi-view setup consists of 6 virtual cameras uniformly distributed between the left and right views, with each camera generating a corresponding warped video sequence. Due to computational constraints of the Zeroscope model, we limit our experiments to 16-frame video sequences. For the denoising process, we adopt the DDPM framework~\cite{ho2020denoising} with $T=1000$ total timesteps and perform denoising in 50 steps (that is, jumps of 20 timesteps per step). Following the RePaint strategy~\cite{lugmayr2022repaint}, we implement an adaptive resampling schedule: During the initial coarse denoising phase (steps 50--25), we apply 8 resampling iterations per step to establish coherent structures in disoccluded regions. In the refinement phase (steps 25--0), we reduce to 4 resamples per step and focus denoising exclusively on the right view for computational efficiency. \textit{2) Spatial video.} Different from the implementation of stereoscopic video, we place 16 cameras on a circular trajectory, with the radius set as $0.07$ meters, surrounding the reference viewpoint and denoise the entire frame matrix to obtain multi-view videos. When optimizing the 4D Gaussian, $\lambda_1$ and $\lambda_2$ are set to 1.0 and 0.15 respectively, and the total number of optimization iterations is 20000. For training efficieny, the $LPIPS$ loss is evoked after 13000 iterations.

\vspace{0.1in}
\noindent\textbf{Baselines.}
We first evaluate our method against two families of approaches: video inpainting and novel view synthesis from a monocular video.
For video inpainting approaches, we produce right views through depth-guided warping identical to our pipeline, then process them with state-of-the-art inpainting methods: ProPainter~\cite{zhou2023propainter} and E2FGVI~\cite{li2022towards}.
For novel view synthesis methods, we compare our results with RoDynRF~\cite{liu2023robust} and DynIBaR~\cite{li2023dynibar}, which optimize scene representations relying on camera poses. To ensure a fair comparison, given the differing 3D scales between their reconstructed scenes and our estimated depth, we align their rendering baseline with ours by matching median foreground disparities in the resulting disparity maps. We also include comparisons with single-image view synthesis approaches, such as SVM~\cite{tucker2020single} and AdaMPI~\cite{han2022single}, which fail to leverage temporal information. 
Next, we are also aware of approaches trained on dedicated datasets that directly produce the right view given the left view, like Deep3D~\cite{xie2016deep3d}.
However, it does not generalize well to generated videos, especially those in non-realistic styles. 
Finally, we compare with the concurrent work Free4D~\cite{liu2025free4d} that generates 4D scene based on the geometric structure from MonST3R~\cite{zhang2024monst3r}.

\begin{figure}[!htb]
    \centering
    \includegraphics[width=1.0\linewidth]{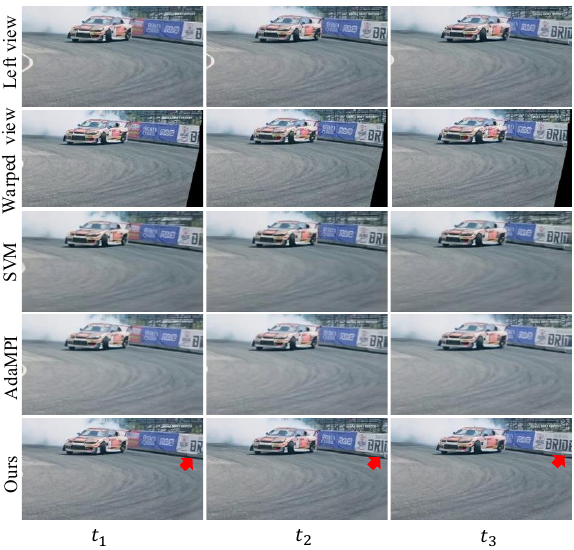}
    \vspace{-0.25in}
    \caption{\textbf{Modeling temporal relationships.} Our method leverages future information to inpaint disoccluded regions. Please note that the generated character ``R'' at time $t_1$ matches the left-view character at time $t_2$. Single-image view synthesis methods (SVM and AdaMPI) produce temporally inconsistent results. }
    \label{fig:stereo_image}
    \vspace{-0.1in}
\end{figure}

\begin{figure}[!htb]
    \centering
    \includegraphics[width=0.98\linewidth]{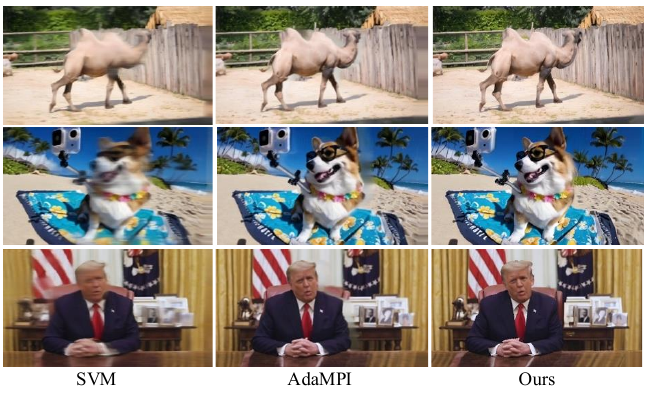}
    \vspace{-0.1in}
    \caption{\textbf{More comparisons with single-image view synthesis.} SVM and AdaMPI tend to produce blurry results in disoccluded regions.}
    \label{fig:more_stereo_image}
    \vspace{-0.1in}
\end{figure}


\begin{figure}[!htb]
    \centering
    \includegraphics[width=0.98\linewidth]{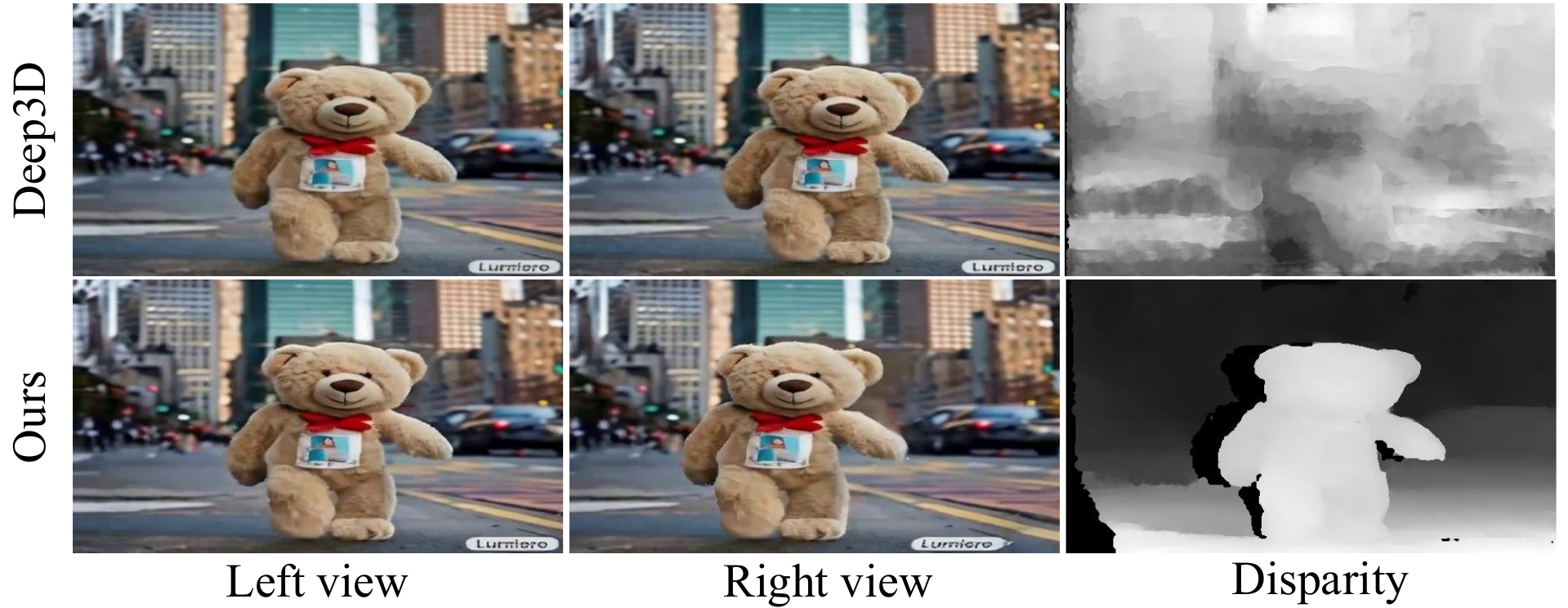}
    \vspace{-0.05in}
    \caption{\textbf{Results of Deep3D.} The predicted disparity map~\cite{Li_2021_ICCV} of Deep3D is blurry, indicating its weak 3D effect. Moreover, Deep3D can only generate fixed binocular videos and does not provide the flexibility of changing the stereo baseline.}
    \label{fig:deep3d}
\end{figure}

\begin{figure*}[!htb]
    \centering
    \includegraphics[width=1.0\linewidth]{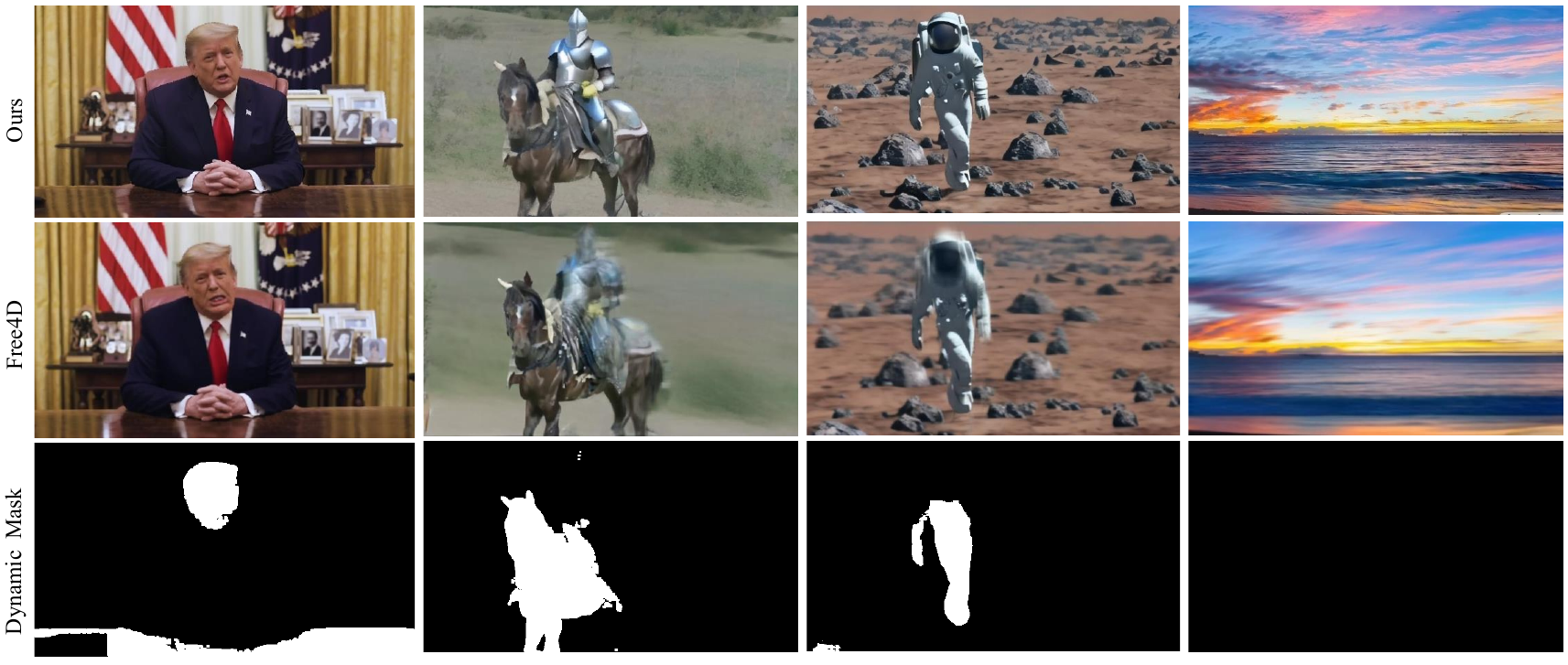}
    \vspace{-0.2in}
    \caption{\textbf{Qualitative comparisons on spatial generation.} We evaluate the spatial generation capability by comparing rendered images from the generated 4D scenes. Our approach outperforms Free4D in terms of image quality and sharpness, especially on scenes with complex motions, such as the running horse and non-rigid water flowing, where Free4D produces blurry results. Additionally, we display the dynamic mask, a key component of Free4D, predicted by MonST3R, which needs further fine-tuning to accurately represent the dynamic regions in a large variety of scenes with complex motions.}
    \label{fig:comparisons_spatial}
\end{figure*}

\begin{table*}[!htb]
    \caption{\textbf{Human perception.} This table reports the results of human perception experiments as mean (std). 
    Our method outperforms all other training-free baselines on all metrics, excluding Deep3D, which is fine-tuned on private stereo videos. Deep3D achieves slightly better video quality and consistency, but significantly sacrifices the stereo effect and overall experience. 
    }
    \vspace{-0.1in}
    \centering
    {
    \begin{tabular}{ccccccc}
    \toprule
    & E2FGVI & ProPainter & RoDynRF & DynIBaR & Deep3D & Ours\\
    \midrule
    Stereo Effect~$\uparrow$  & 4.79 (1.08) & \underline{4.81} (1.13) & 2.97 (1.34) & 1.86 (1.25) & 2.27 (1.40) & \textbf{5.24} (0.94) \\
    Temporal Consistency~$\uparrow$  & 4.74 (1.33) & 4.74 (1.22) & 3.35 (1.66) & 1.89 (1.33) & \textbf{5.46} (1.20) & \underline{5.15} (1.22) \\
    Image Quality~$\uparrow$  & 4.42 (1.27) & 4.38 (1.28) & 2.84 (1.60) & 1.67 (1.07) & \textbf{5.27} (1.31) & \underline{5.12} (1.33) \\
    Overall Experience~$\uparrow$  & \underline{4.67} (1.04) & 4.66 (1.09) & 2.92 (1.43) & 1.72 (1.06) & 3.88 (0.96) & \textbf{5.35} (0.99)\\
    \bottomrule 
    \end{tabular}}
    \label{tab:human_perception_1}
\end{table*}

\begin{table*}[!htb]
    \caption{\textbf{Quantitative comparisons on stereoscopic generation}. We show the semantic consistency using CLIP feature similarity~\cite{hessel2021clipscore} between the left and right views. Additionally, the quality of generated videos is measured by aesthetic score~\cite{schuhmann2022laion}, DOVER~\cite{wu2023exploring}, and FVD~\cite{unterthiner2019fvd}. Our method outperforms previous methods.}
    \vspace{-0.1in}
    \label{tab:quanti_results}
    \centering
    {
    \begin{tabular}{cccccccc}
    \toprule
    Method & E2FGVI & ProPainter & RoDynRF & DynIBaR & SVM & AdaMPI & Ours\\
    \midrule
    CLIP~$\uparrow$ & 94.34 & 95.29 & 96.03 & 93.24 & 92.11 & 93.61 & \textbf{96.44}\\
    Aesthetic~$\uparrow$ & 5.06 & 5.07 & 4.97 & 4.66 & 4.87 & 4.78 & \textbf{5.27}\\
    DOVER~$\uparrow$ & 0.547 & 0.535 & 0.352 & 0.365 & 0.215 & 0.245 & \textbf{0.584}\\
    FVD~$\downarrow$ & 638 & 606 & 727 & 1208 & 784 & 718 & \textbf{599}\\
    \bottomrule
    \end{tabular}}
\end{table*}

\begin{table}[!htb]
    \caption{{\textbf{Ablation studies}. We study the efficacy of proposed components, including frame matrix, disocclusion boundary re-injection, data processing, and the number of cameras used.} }
    \vspace{-0.1in}
    \label{tab:ablation_study}
    \centering
    {
    \begin{tabular}{cccccc}
    \toprule
     & -FM & -DBR & -DP & Ours (4 cams) & Ours\\
    \midrule
    CLIP~$\uparrow$ & 95.81 & 95.60 & 95.68 & \textbf{96.50} & 96.44\\
    Aesthetic~$\uparrow$ & 5.25 & 5.18 & 5.25 & 5.26 & \textbf{5.27}\\
    DOVER~$\uparrow$ & 0.565 & 0.560 & 0.576 & 0.565 &\textbf{0.584}\\
    FVD~$\downarrow$ & 614 & 699 & 733& \textbf{579} & 599\\
    \bottomrule
    \end{tabular}}
\end{table}

\begin{table}
    \caption{\textbf{Quantitative comparisons on spatial generation}. We render videos from established 4D scenes and measure the quality of rendered videos using the aesthetic score, Dover, and FVD. Our method outperforms Free4D.}
    \vspace{-0.1in}
    \label{tab:quantitative_spatial}
    \centering
     {
    \begin{tabular}{cccc}
    \toprule
    Method & Aesthetic~$\uparrow$ & DOVER~$\uparrow$ & FVD~$\downarrow$ \\
    \midrule
    Free4D & 4.97 & 0.371 & 1129\\
    Ours & \textbf{5.18} & \textbf{0.567} & \textbf{518}\\
    \bottomrule
    \end{tabular}}

    \centering

\end{table}

\subsection{Qualitative Results}
\noindent\textbf{Frames in Generated Frame Matrix.} In Fig.~\ref{fig:multi_view_traj}, we sample frames across different camera viewpoints and timestamps in the inpainted frame matrix. Each row represents a video with camera motions at a specific timestamp; the corresponding camera positions are displayed in the top left corner. Each column contains frames at different timestamps, as observed from a camera viewpoint. The results exhibit spatial-temporal coherence in both the foreground and background components.  

\vspace{0.1in}
\noindent\textbf{Qualitative Comparisons on Stereoscopic Generation.} 
We present qualitative comparisons in Fig.~\ref{fig:comparisons}. Existing video inpainting methods exhibit a common limitation: the generated content in disoccluded regions (\textit{e.g.}, the knight's arm, horse's tail, corgi's face, and turtle's head) often appears blurry, likely due to training on limited datasets. On the other hand, novel view synthesis methods struggle with unstable camera pose estimation (\textit{e.g.}, DynIBaR's failure on certain videos). While these methods excel at reconstructing visible content from monocular video, they are typically poor at synthesizing novel contents in the disoccluded regions that are not observed in any frames (\textit{e.g.},  ghost effect near the boundary in the RoDynRF result on the corgi example).
Our approach leverages the generative power of video diffusion models trained on large-scale datasets and eliminates the need for input camera poses. As demonstrated in the last row of Fig.~\ref{fig:comparisons}, our method generates high-quality content in various scenarios and consistently exceeds baselines.

Without using input camera poses, single-image view synthesis methods can produce stereoscopic videos by processing each frame individually. We compare our approach with SVM~\cite{tucker2020single} and AdaMPI~\cite{han2022single}, presenting results at different timestamps in Fig.~\ref{fig:stereo_image}. At a specific timestamp (e.g., $t_1$), both SVM and AdaMPI produce blurry artifacts near the character ``B'', whereas our method reconstructs a sharp character ``R''. This enhancement stems from our framework's ability to aggregate multi-frame information (e.g., leveraging the left-view frame at $t_2$). Furthermore, AdaMPI and SVM exhibit temporal inconsistencies due to their per-frame processing paradigm; for example, the region adjacent to ``B'' varies between $t_1$ and $t_2$. In contrast, our video generation model enforces temporal coherence, ensuring alignment of synthesized characters with the left-view sequence. Additional comparative results are provided in Fig.~\ref{fig:more_stereo_image}.

Furthermore, we compare our method with Deep3D~\cite{xie2016deep3d}, which is trained on private stereoscopic videos. In Fig.~\ref{fig:deep3d}, we visualize 3D effects by predicting the disparity map from generated stereoscopic pairs using STTR~\cite{Li_2021_ICCV}. The vague disparity map of Deep3D in the third column demonstrates its weak stereo effects, while our method displays a clear order of foreground and background content with smooth depth changes. Moreover, Deep3D does not support manually modifying the disparity map or changing the stereo baseline to achieve different stereo effects. 

\begin{figure*}[!htb]
    \centering
    \includegraphics[width=0.98\linewidth]{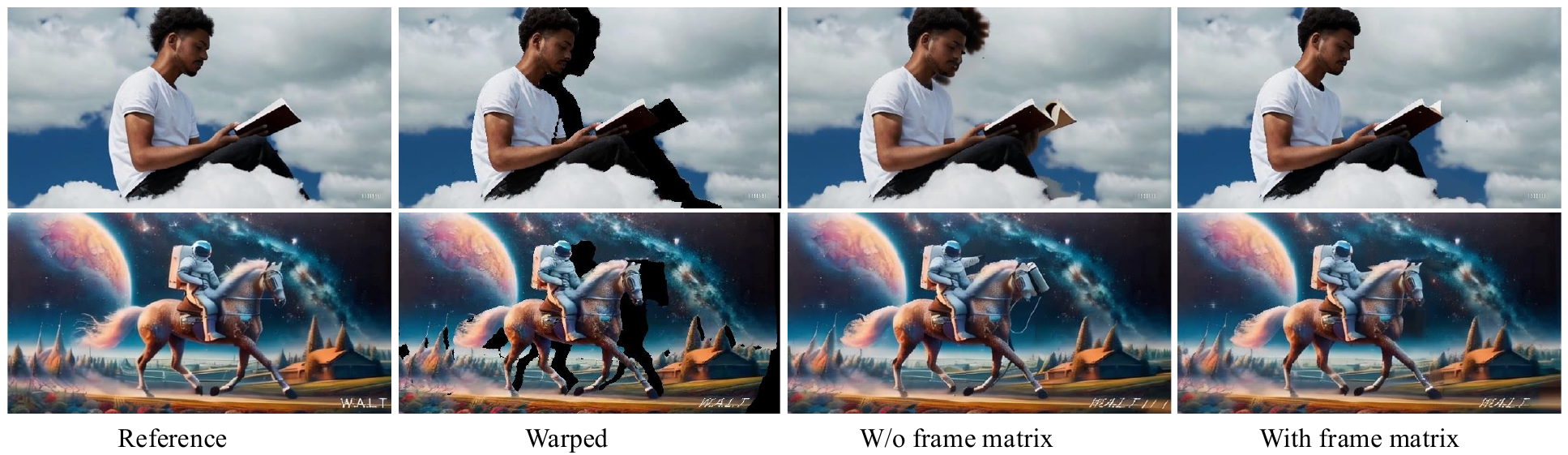}
    \vspace{-0.1in}
    \caption{\textbf{Semantic coherence across views.} Warp the reference frame into the target view and set the occluded areas to black. Without using the frame matrix, the generated content does not match the reference image, such as the book and the horse's face. With the frame matrix, the inpainted content is more semantically reasonable across views.}
    \label{fig:sematic_matching}
    \vspace{-0.1in}
\end{figure*}

\begin{figure*}[!htb]
    \centering
    \includegraphics[width=0.98\linewidth]{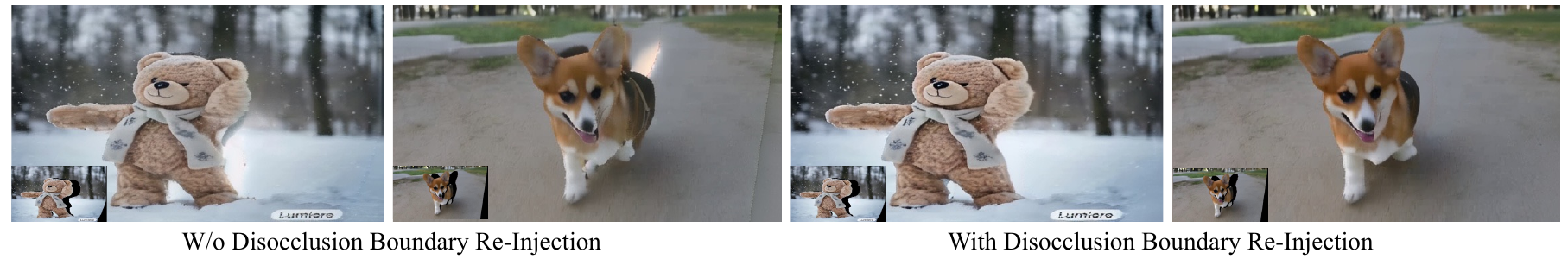}
    \vspace{-0.1in}
    \caption{\textbf{Disocclusion boundary re-injection.} Without disocclusion boundary re-injection for latent feature updating, the inpainted images usually contain artifacts. The warped image is shown at the bottom left.}
    \label{fig:update_feature}
    \vspace{-0.1in}
\end{figure*}

\begin{figure*}[!htb]
    \centering
    \includegraphics[width=0.98\linewidth]{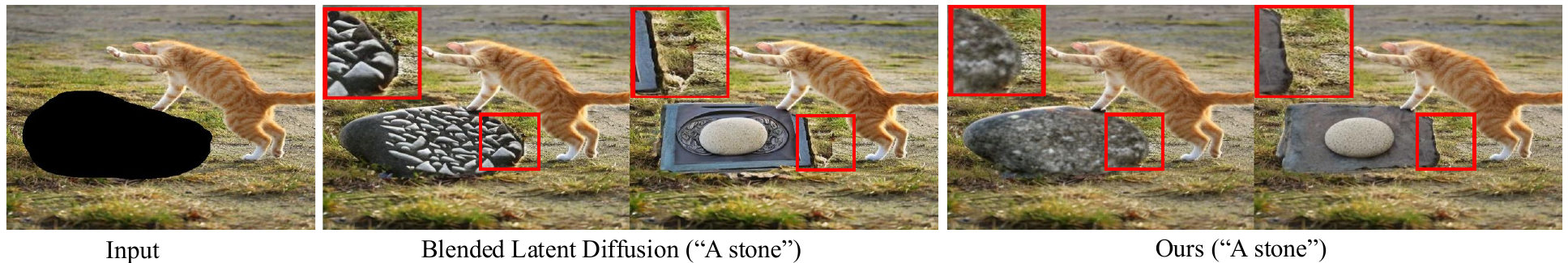}
    \vspace{-0.05in}
    \caption{\textbf{Disocclusion boundary re-injection in image inpainting.} Our approach (blended latent diffusion + disocclusion boundary re-injection) provides a smoother transition between original and inpainted content, such as the grassland.}
    \label{fig:image_inpainting}
    \vspace{-0.1in}
\end{figure*}

\begin{figure*}[!htb]
    \centering
    \includegraphics[width=0.98\linewidth]{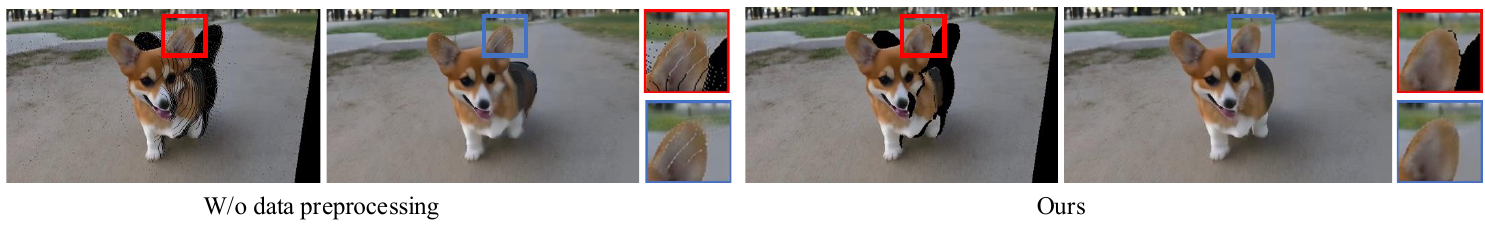}
    \vspace{-0.1in}
    \caption{\textbf{Isolated points and cracks.} Left: without handling isolated points and entangled foreground and background (the gray road is observed through cracks) in warped images, these artifacts remain in the final results. Right: our results are clean.}
    \label{fig:ab_data_preprocessing}
    \vspace{-0.1in}
\end{figure*}

\begin{figure*}[!htb]
    \centering
    \includegraphics[width=0.98\linewidth]{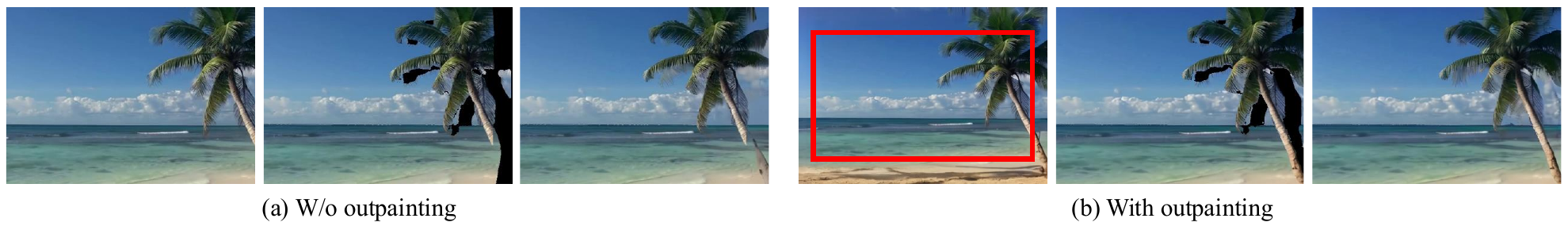}
    \vspace{-0.1in}
    \caption{\textbf{Partially observed object.} (a) The coconut tree is incomplete in the video. Using depth-based warping causes wrong occlusion relationships, where the background sky occupies regions belonging to the trunk. This problem cannot be fixed by video inpainting. (b) We outpaint the input video (regions outside the red box), thus providing a more complete coconut tree for correct warping and inpainting.}
    \label{fig:incomplete}
\end{figure*}

\begin{figure*}[!htb]
    \centering
    \includegraphics[width=0.99\linewidth]{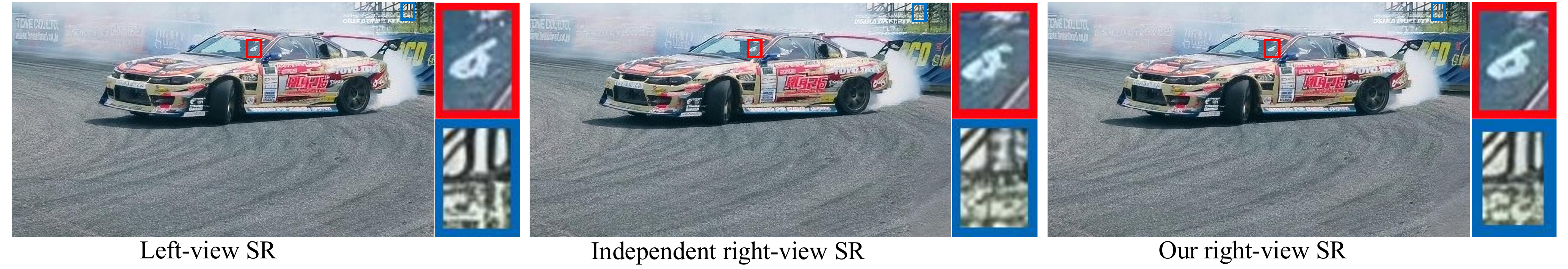}
    \vspace{-0.15in}
    \caption{\textbf{Stereoscopic super resolution.} Independent upsampling of left-view and right-view videos often leads to binocular inconsistency, as demonstrated in the first two columns. Our method maintains cross-view consistency by explicitly coupling the binocular information during the upsampling process.}
    \label{fig:SR}
    \vspace{-0.1in}
\end{figure*}

\begin{figure*}[!htb]
    \centering
    \includegraphics[width=1.0\linewidth]{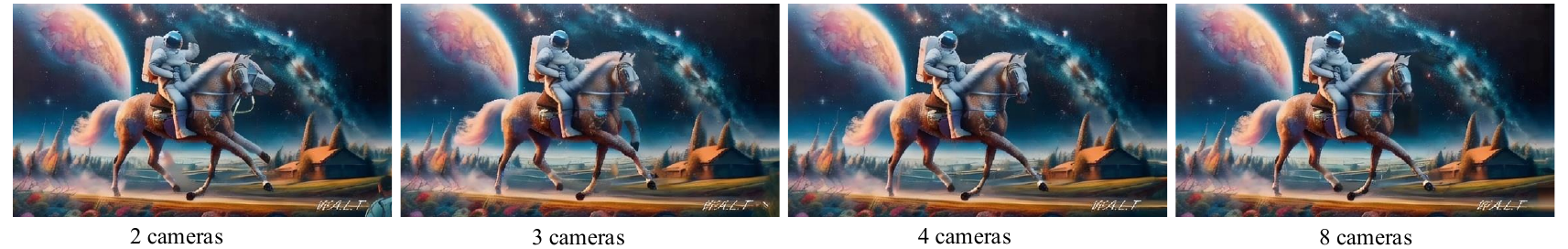}
    \vspace{-0.3in}
    \caption{\textbf{Different number of cameras used.} Artifacts arise when the number of cameras between left and right views is too small to exceed the capability of handling abrupt changes of video generation models.}
    \label{fig:num_cameras}
    \vspace{-0.1in}
\end{figure*}

\begin{figure*}[!htb]
    \centering
    \includegraphics[width=0.98\linewidth]{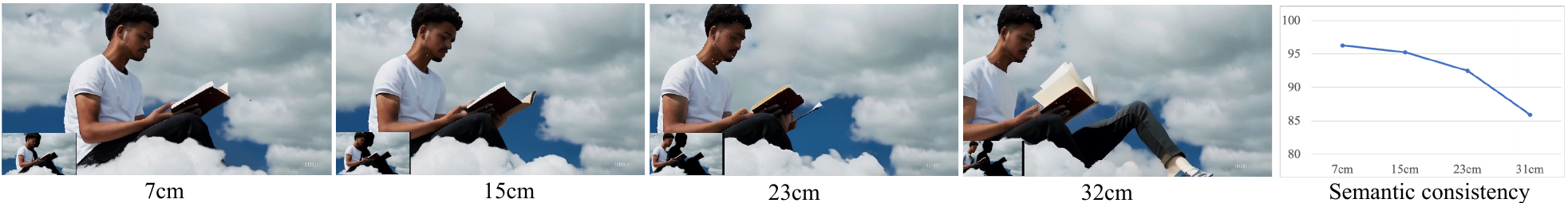}
    \vspace{-0.1in}
    \caption{\textbf{Different stereo baselines.} Unnatural artifacts begin to appear as the baseline expands. Our method performs well for stereoscopic video generation where the baseline is usually less than 7cm.}
    \label{fig:stereo_baslie}
    \vspace{-0.1in}
\end{figure*}

\begin{figure}[!htb]
    \centering
    \includegraphics[width=1.0\linewidth]{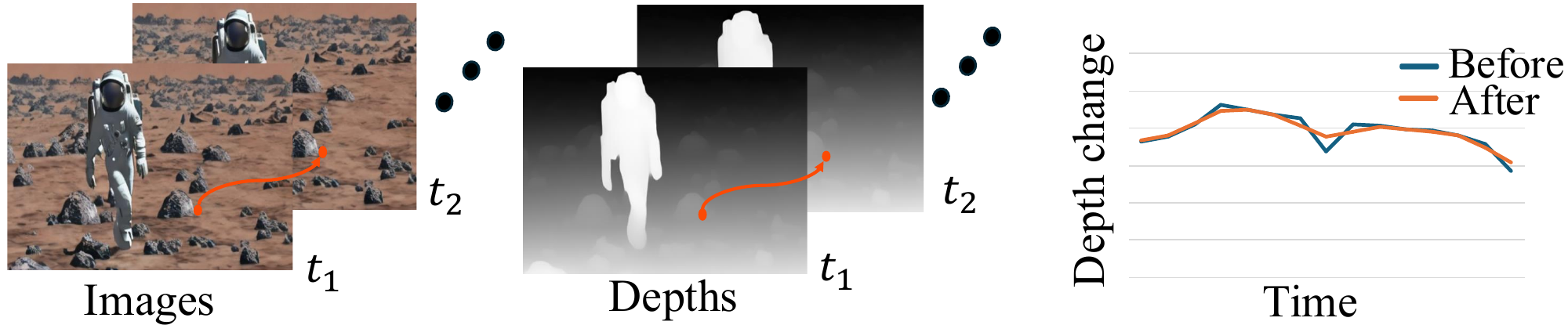}
    \vspace{-0.2in}
    \caption{\textbf{Temporal depth smoothing.} We track a pixel's depth values across frames and visualize the depth changes. By applying smoothing operations, the depth changes more smoothly.}
    \label{fig:depth_changes}
    \vspace{-0.1in}
\end{figure}

\begin{figure}[!htb]
    \centering
    \includegraphics[width=1.0\linewidth]{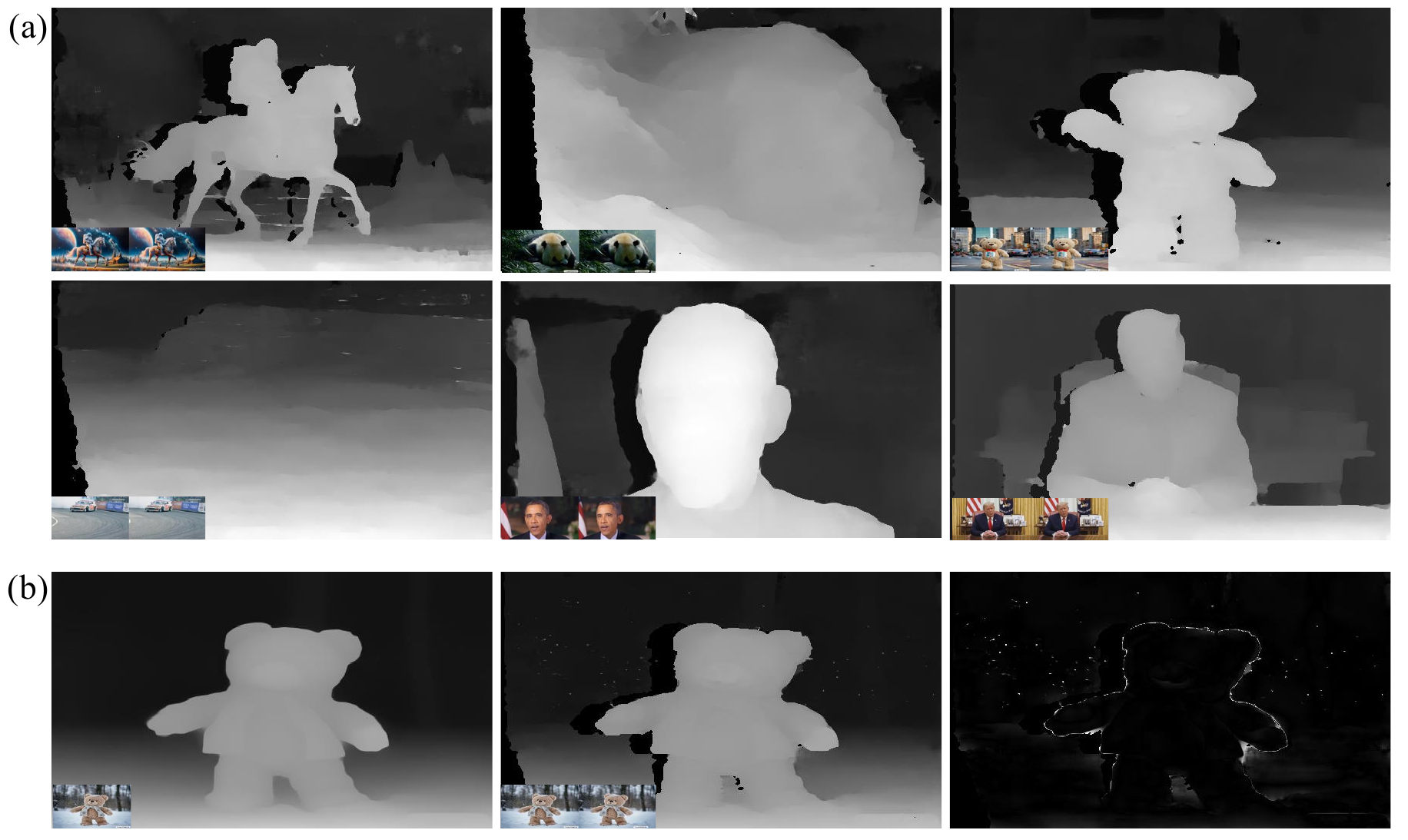}
    \vspace{-0.2in}
    \caption{\textbf{3D effects visualization.} (a) The generated stereo results are used for predicting disparity maps~\cite{Li_2021_ICCV}. The clear distinction between foreground and background content indicates reasonable 3D effects. (b) Disparity maps are estimated from monocular and binocular images. The discrepancy map on the right side demonstrates their consistency.}
    \label{fig:stereo_exp}
    \vspace{-0.1in}
\end{figure}

\begin{figure}[!htb]
    \centering
    \includegraphics[width=1.0\linewidth]{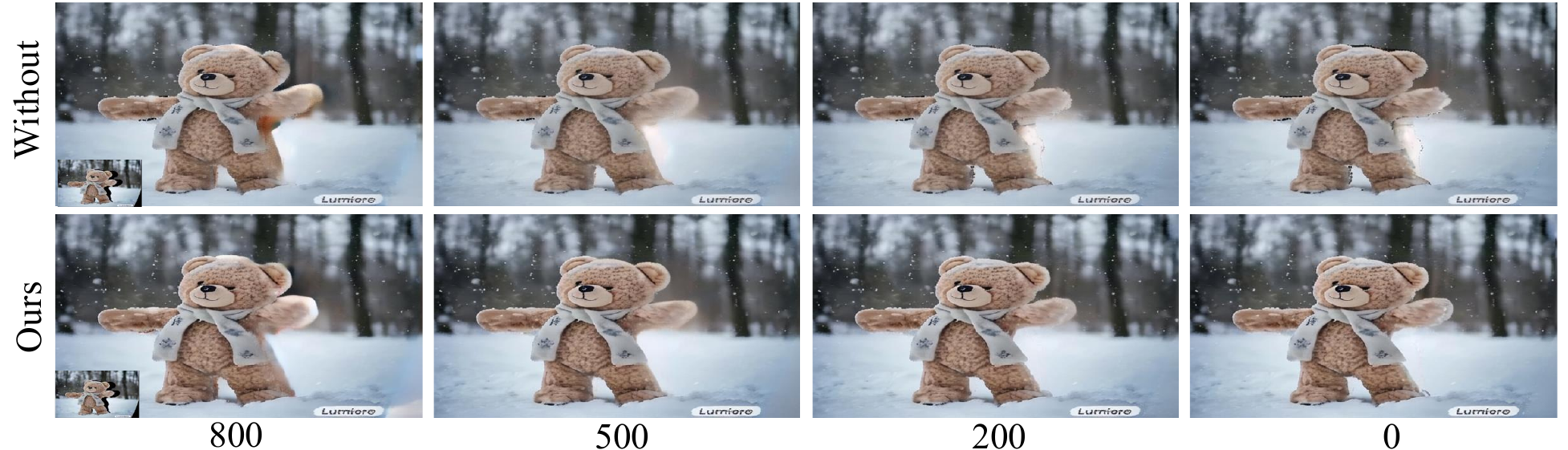}
    \vspace{-0.2in}
    \caption{\textbf{Results at different denoising timesteps.} Without the use of disocclusion boundary re-injection, artifacts persist throughout the denoising process. In contrast, our method gradually fills in disoccluded regions with harmonious content.}
    \label{fig:step_by_step}
    \vspace{-0.1in}
\end{figure}

\begin{figure}[!htb]
    \centering
    \includegraphics[width=1.0\linewidth]{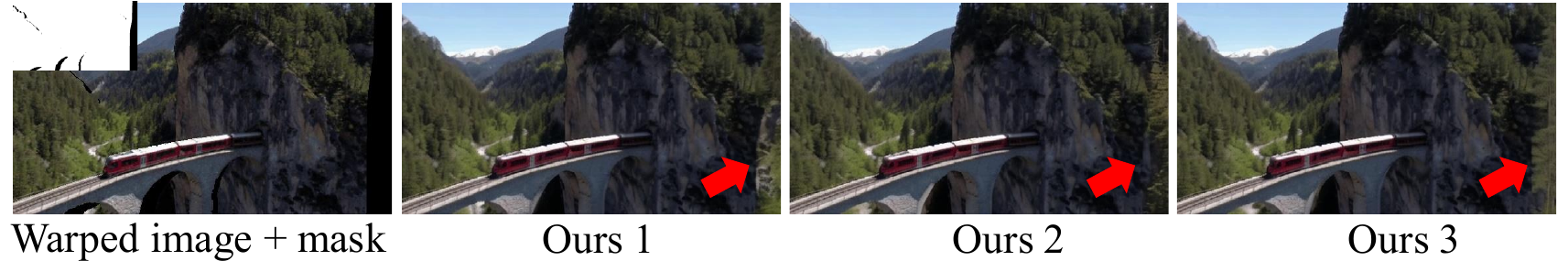}
    \vspace{-0.2in}
    \caption{\textbf{Diverse outputs.} The disoccluded regions can be generated with different content.}
    \label{fig:diverse_outputs}
    \vspace{-0.1in}
\end{figure}

\begin{figure*}[!htb]
    \centering
    \includegraphics[width=0.99\linewidth]{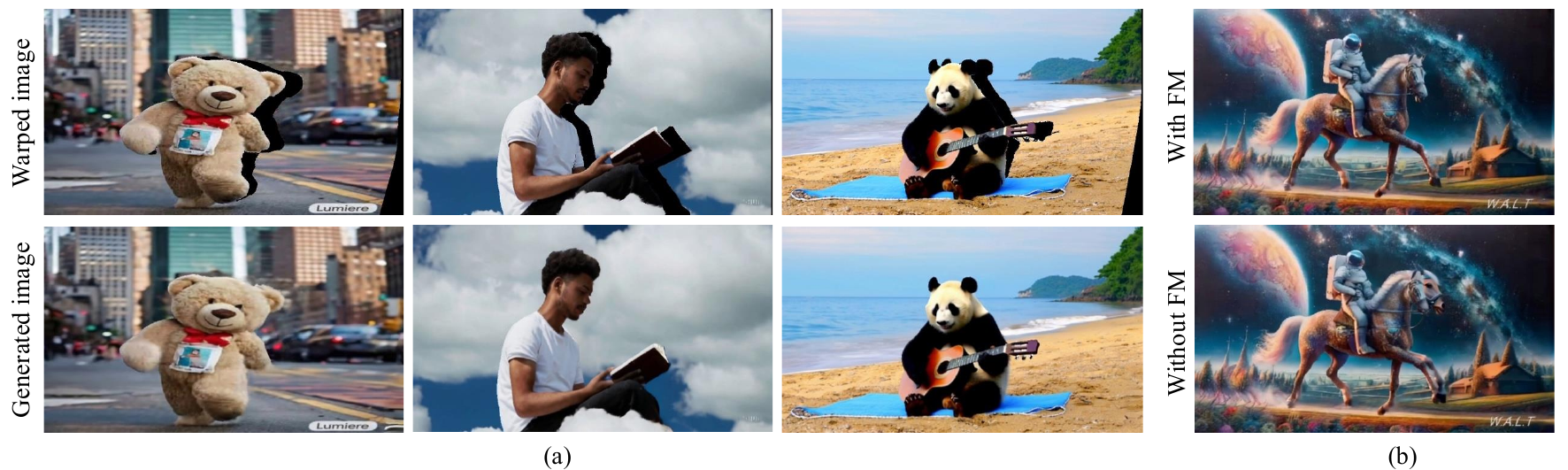}
    \vspace{-0.1in}
    \caption{{\textbf{Results using Wan 2.1 as the backbone}. (a) The warped images and the corresponding generated images. (b) Without leveraging spatial-direction denoising, results suffer from semantic inconsistency.}}
    \label{fig:wan}
    \vspace{-0.1in}
\end{figure*}

\vspace{0.1in}
\noindent\textbf{Qualitative Comparisons on Spatial Generation.} In Fig.~\ref{fig:comparisons_spatial}, we display rendered images from optimized 4D scenes. Our method outperforms Free4D~\cite{liu2025free4d} in terms of rendering quality and sharpness. Since Free4D employs MonST3R's results as the geometry initialization, which reconstructs dynamic and static content in the 4D scene separately, we show the output dynamic masks (Fig.~\ref{fig:comparisons_spatial}, third row), where white represents dynamic regions. From these dynamic masks, we observed that precisely segmenting the dynamic regions remains difficult, especially in scenes with complex motions, such as the fast-moving horse and the non-rigid motion of water flowing. Consequently, the failure to separate dynamic and static regions results in blurred reconstruction in Free4D. In contrast, our method circumvents predicting dynamic regions; instead, we treat each Gaussian primitive as a potential dynamic entity and learn a time-dependent deformation offset. Moreover, Free4D's performance degrades when the input video contains camera motions. For example, the case of Trump with a fixed camera viewpoint performs better than other cases. This is because camera motions exaggerate task difficulty, and the temporal replacement operation in Free4D is not suitable for input videos with viewpoint changes. Video comparisons are displayed on our project page, where our method showcases temporally consistent results.


\subsection{Quantitative Results}
This section presents comprehensive quantitative comparisons with state-of-the-art baseline methods. Our evaluation framework combines both subjective and objective assessment metrics: (1) a rigorously designed user study evaluating multiple quality dimensions of the generated stereoscopic videos, (2) an objective metric based on the CLIP model that quantifies semantic alignment between left and right views, and (3) three established metrics measuring the quality of generated videos. 

\vspace{0.08in}
\noindent\textbf{Human Perception.}
To evaluate perceived visual quality, we conducted a controlled user study involving 20 participants (9 female, age $\mu=33, \sigma=6.2$). Using a VR headset, each participant assessed five randomly selected videos (from a pool of 20) generated by all five methods. The evaluation covered four key dimensions: stereo effect, temporal consistency, image quality, and overall experience, measured using a 7-point Likert scale~\cite{likert1932technique}.
A total of 435 evaluations (DynIBaR failed to generate 13 videos) were counterbalanced and randomly shuffled. We also included a training session to eliminate novelty effects. Results are summarized in Table~\ref{tab:human_perception_1}, our method outperforms other view synthesis and video inpainting baselines in measured metrics. Deep3D achieves good video quality and consistency via slightly modifying the input reference video, which significantly sacrifices stereo effects (demonstrated by Fig.~\ref{fig:deep3d}) and overall experience.

\vspace{0.08in}
\noindent\textbf{Semantic Consistency.} 
To further evaluate view consistency, we introduce a semantic alignment metric based on pre-trained CLIP features~\cite{radford2021learning}. For each stereoscopic video, we extract feature embeddings from both left and right views, then compute their similarity using the distance metric proposed in~\cite{taited2023CLIPScore}. As demonstrated in Table~\ref{tab:quanti_results}, our method achieves superior semantic consistency with a score of 96.44, outperforming baseline approaches.

\vspace{0.08in}
\noindent\textbf{Video Quality Assessment.} 
We evaluate video quality using three established metrics aligned with human judgment: (1) the aesthetic score~$\uparrow$~\cite{schuhmann2022laion}, (2) DOVER~$\uparrow$~\cite{wu2023exploring}, and (3) FVD~$\downarrow$~\cite{unterthiner2019fvd}. In Table~\ref{tab:quanti_results}, our approach in stereoscopic generation achieves the best performance (5.27, 0.584, and 599), aligning with the conclusion from the user study experiment in Table~\ref{tab:human_perception_1}. When evaluating spatial generation, we randomly sample six viewpoints and render videos from the optimized 4D scenes. We report the measured video quality in Table~\ref{tab:quantitative_spatial}, where our approach outperforms Free4D ($5.18~vs.~4.97,~ 0.567~vs.~0.371,~\text{and}~ 518~vs.~1129$).

\subsection{Ablation Studies}
\noindent\textbf{Frame Matrix.}
Fig.~\ref{fig:sematic_matching} demonstrates the critical role of the frame matrix in maintaining semantic consistency between left and right views. When the frame matrix is disabled, the warped images' disoccluded regions (e.g., the man's hair and horse's head) exhibit inconsistent content despite the diffusion model's strong generative capabilities. {This is quantitatively confirmed in Table~\ref{tab:ablation_study}, where disabling the frame matrix reduces the CLIP Score from 96.44 to 95.81.} Thanks to the frame matrix that provides essential constraints from neighboring frames, ensuring both foreground and background content in disoccluded regions remain plausible and coherent.

\vspace{0.1in}
\noindent\textbf{Disocclusion Boundary Re-Injection.}
Fig.~\ref{fig:update_feature} demonstrates the importance of updating unoccluded latent features for achieving high-quality results. When this update step is omitted, the disoccluded region is inpainted with unnatural textures that cannot blend well with the surrounding warped content. In contrast, our approach with feature updating produces seamless content integration, as evidenced by both visual results and quantitative metrics. {Specifically, Table~\ref{tab:ablation_study} shows that disabling feature updates leads to a measurable degradation in performance, with the aesthetic score dropping from 5.27 to 5.18.} Additionally, this boundary re-injection design demonstrates broader applicability beyond video inpainting, as evidenced by its effectiveness in improving existing image inpainting approaches. Fig.~\ref{fig:image_inpainting} shows consistent quality improvements when applied to methods like blended latent diffusion~\cite{avrahami2023blended}, where the technique helps reduce artifacts near inpainting boundaries.

\begin{table}[!htb]
    \caption{{\textbf{Performance on Wan 2.1}. We report quantitative results on stereo and spatial video generation using Wan 2.1 as the backbone}}.
    \vspace{-0.1in}
    \label{tab:wan2.1}
    \centering
     {
    \begin{tabular}{ccccc}
    \toprule
    Task & Aesthetic~$\uparrow$ & DOVER~$\uparrow$ & FVD~$\downarrow$ & CLIP~$\uparrow$\\
    \midrule
    Stereo generation & 5.23 & 0.566 & 595 & 96.48\\
    Spatial generation & 5.16 & 0.558 & 488 & $-$\\
    \bottomrule
    \end{tabular}}
    \centering

\end{table}

\vspace{0.1in}
\noindent\textbf{Data Processing.}
(1) \textit{Handling Isolated Pixels and Cracks.} Fig.~\ref{fig:ab_data_preprocessing} (left) reveals significant artifacts in the warped images, including isolated points and visible cracks where the foreground ear improperly blends with the gray road background. Crucially, these artifacts persist in the final generated output without our proposed processing. After implementing our warping framework described in Section~\ref{sec:video_warp}, the results demonstrate marked improvement, as shown in Fig.~\ref{fig:ab_data_preprocessing} (right). Our approach successfully eliminates these warping artifacts while maintaining both foreground and background structural integrity. {In Table~\ref{tab:ablation_study}, data preprocessing has influence on all evaluation metrics, suggesting that artifacts such as cracks and noisy points can disrupt the integrity of the generation process.} (2) \textit{Handling Partially Observed Objects.} Fig.~\ref{fig:incomplete} (a) shows a partially observed coconut tree and its depth-based warping results. Since regions belonging to the unobserved trunk are occupied by the background sky, these regions cannot be modified by the following video inpainting, leading to an incomplete coconut tree in the final output. To address this, we propose to outpaint the video (regions outside the red box), which provides a more complete coconut tree for warping and inpainting, thereby preserving correct occlusion relationships, as shown in Fig.~\ref{fig:incomplete} (b). 

\vspace{0.1in}
\noindent\textbf{Stereoscopic Video Super Resolution.}
In Fig.~\ref{fig:SR}, we present our video super-resolution results. While independent upsampling of left and right views fails to maintain cross-view consistency (evident in the zoomed-in regions), our method preserves stereo coherence by explicitly establishing inter-view connections during the upsampling process.

\vspace{0.1in}
\noindent\textbf{Number of Cameras Used.}
When generating stereoscopic videos, we gradually reduce the number of cameras between the left and right views and show results in Fig.~\ref{fig:num_cameras}. Artifacts (e.g., the fifth leg) tend to arise when the number of cameras is fewer than four in this horse case, as a limited number of cameras can lead to rapid changes between video frames that exceed the capabilities of current video generation models. {Table ~\ref{tab:ablation_study} quantitatively demonstrates that the number of cameras can be reduced to some extent without significantly affecting final results.}

\vspace{0.1in}
\noindent\textbf{Temporal Depth Smoothing.}
In Fig.~\ref{fig:depth_changes}, we track a pixel across video frames and display its depth changes with and without using our depth smoothing operation. From Fig.~\ref{fig:depth_changes} (right), our approach effectively stabilizes depth changes across frames.

\vspace{0.1in}
\noindent\textbf{Different Stereo Baselines.}
Fig.~\ref{fig:stereo_baslie} demonstrates the impact of stereo baseline on inpainting difficulty and output quality.  The performance degrades with the increasing baseline, and our method maintains robustness up to approximately 20cm (with scene depth normalized to 1.0--10.0 meters). This operational range comfortably accommodates typical viewing scenarios, as it substantially exceeds the average human inter-pupillary distance of 5--7cm. The results confirm our method's practical suitability for generating stereoscopic content across realistic viewing conditions.



\vspace{0.1in}
\noindent\textbf{Stereo Effects Visualization.}
To further verify the quality of stereo effects, we compute more disparity maps from our generated stereo videos utilizing STTR~\cite{Li_2021_ICCV}. Fig.~\ref{fig:stereo_exp} (a) demonstrates that our results produce sharp disparity maps with clear foreground-background separation, confirming plausible 3D perception effects. The method achieves this via maintaining strong consistency with the input monocular depth - a natural consequence of our warping-based approach and the mask-aware diffusion model's tendency to preserve known pixels. This is validated in Fig.~\ref{fig:stereo_exp} (b), comparing monocular (first column) and binocular (second column) estimated disparity maps. The small average difference of 0.63 pixels (last column) indicates excellent geometry preservation.

\vspace{0.1in}
\noindent\textbf{Intermediate Denoising Inpainting Results.} 
In Fig.~\ref{fig:step_by_step}, we show results at different denoising time steps to enhance the understanding of the denoising inpainting process. Without the disocclusion boundary re-injection, noticeable artifacts gradually appear during the denoising process. In contrast, our method can gradually fill in disoccluded regions with harmonious content.

\vspace{0.1in}
\noindent\textbf{Diverse Outputs.} As illustrated in Fig.~\ref{fig:diverse_outputs}, the output is non-unique since disoccluded regions can be plausibly synthesized with varying content, such as cliffs or trees.

\subsection{Transferability} 
We conduct experiments on the Wan2.1-T2V-1.3B~\cite{wan2025} video generation model to show the transferability of the proposed method. A key architectural difference is that Zeroscope only compresses spatial resolution in the latent space (e.g., 1/8), whereas newer video generators such as Wan 2.1 additionally compress the temporal dimension (e.g., 1/4). This temporal compression breaks the assumption in our original formulation that each latent feature map corresponds to a distinct frame, making direct sampling along spatial directions in latent space meaningless. In other words, Wan 2.1’s latent representation is no longer frame-aligned, which makes a naive plug-in of our frame matrix (FM) denoising strategy infeasible. To resolve this, we adapt the alternating scheme as follows: we first denoise the noisy latent-space frame matrix (FM) along one direction and decode it into an RGB-space FM; we then sample along the alternated direction in RGB space, re-encode the sampled frames, re-inject noise, and denoise again. Operationally, we use Wan 2.1 for denoising along the temporal direction and retain Zeroscope for the spatial direction, because Wan 2.1 is unstable in generating very short clips (e.g., 9 frames). In practice, the number of virtual cameras is small, so we need a model that provides strong spatial priors under sparse-view situations; empirically, Zeroscope currently performs better in this role than Wan 2.1. We clarify that the hybrid design is a pragmatic choice reflecting current model behavior, and it also highlights the flexibility of our framework to combine complementary models.
During alternating denoising, the noise level is annealed progressively (e.g., $1000\rightarrow850\rightarrow600\rightarrow400$).

In Fig.~\ref{fig:wan} (a), we display warped images and the generated target-view images using Wan 2.1. Similarly, without applying alternate denoising, the generated content of Wan 2.1 suffers from semantic inconsistency in Fig.~\ref{fig:wan} (b), which demonstrates the efficacy of spatial priors. Quantitatively, in Table~\ref{tab:wan2.1}, leveraging Wan 2.1 yields similar CLIP and FVD scores, but a slightly lower DOVER value (by 0.18) compared to Zeroscope's results. We speculate that this is because Wan 2.1 handles a more challenging denoising inpainting process, as the disoccluded regions of 480p videos are $2.25\times$ larger than those of 360p videos, and the latent space is highly compressed due to temporal compression. Additionally, Wan 2.1 offers enhanced practicality by supporting the processing of longer, higher-resolution video inputs (e.g., 81 frames at 480p). For a comprehensive evaluation, we recommend reviewing the video results.

\begin{figure}[!htb]
    \centering
    \includegraphics[width=0.95\linewidth]{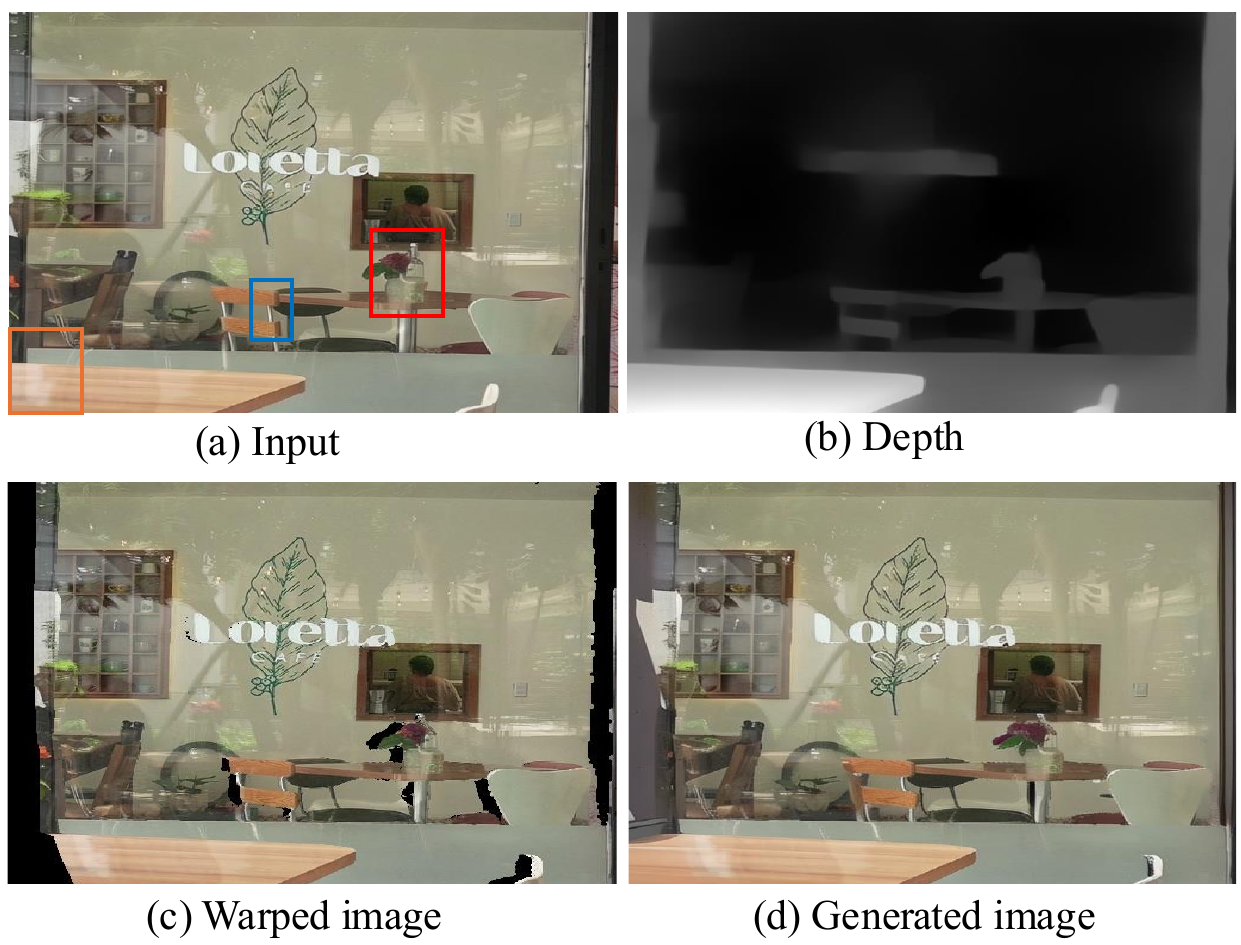}
    \vspace{-0.2in}
    \caption{{\textbf{Failure case.} (a) A scene contains reflective surfaces, thin structures, and transparent objects. (b) Depth estimation fails to predict accurate depths in these regions. (c) Distortions appear after the warping operation. (d) Artifacts are retained in the final generated results.}}
    \label{fig:failure_case}
    \vspace{-0.1in}
\end{figure}

\section{Conclusion}
In this work, we bring the potential and open challenges of immersive content creation using video generation models into focus. Specifically, we present a comprehensive system for stereoscopic and spatial video generation that combines a video diffusion model with our novel \textit{frame matrix} inpainting scheme and pose-free 4D optimization. As video generation technology advances rapidly, our work addresses the critical gap between monocular and immersive video synthesis capabilities. Specifically, we demonstrate that our \textit{frame matrix} formulation and 4D optimization achieve state-of-the-art performance in generative stereoscopic and spatial videos and can be easily integrated into existing video diffusion frameworks.

\vspace{0.1in}\noindent\textbf{Limitations and Discussions.} 
While our results demonstrate the feasibility of generating 3D immersive videos using pre-trained monocular video diffusion models, several challenges remain. First, our study is limited to short video clips due to the inherent constraints of current video diffusion architectures and memory limitations, which typically generate only a few seconds of footage. Likewise, similar issues persist in 4D optimization, where modeling long-range scene motion remains challenging. This limitation could perhaps be addressed by sequentially generating overlapping short clips. 
Second, our method depends on depth estimation models, which can struggle with reflective surfaces, thin structures, transparent objects, and object boundaries (Fig.~\ref{fig:failure_case} (a),( b)). Inaccuracies in these depth estimates are propagated into the warping process, causing distortions that persist in the final output (Fig.~\ref{fig:failure_case} (c),(d)). As our framework inherently benefits from progress in depth estimation, these limitations are expected to be mitigated by improved models. Furthermore, integrating recent advances in video depth estimation (e.g., DepthCrafter~\cite{hu2025depthcrafter}) and metric depth estimation (e.g., Moge~\cite{wang2025moge}) is a viable path towards achieving temporally consistent and scale-accurate results.
Third,  our method is originally designed for video generation models (e.g., Zeroscope) without temporal compression, where each latent map corresponds to a frame. To adapt to modern architectures with temporally compressed latent (e.g., Wan 2.1), we introduce a simple decode-then-encode scheme, which enables generating higher-resolution and longer stereoscopic/spatial videos, albeit with a slightly lower DOVER score. We believe future work should focus on better exploiting the highly compressed latent space (e.g., more principled mask design rather than naive resizing) and improving the stability of modern models on very short clips for providing strong spatial priors under sparse-view settings.
Fourth, our method currently does not support large-viewpoint changes, such as extreme back-to-front motion. To address this, it is worth exploring enhancing the internal geometric priors of video generation models by fine-tuning on multi-view video data, as well as developing a multi-stage strategy that alternates small-step warping/inpainting with iterative state updates.
Fifth, our method introduces additional runtime overhead, exacerbated by the high computational cost of video generation models, leading to a lengthy generation process (70 minutes for a $16\times16$ frame matrix on a single 3090 GPU). To alleviate this, future works can explore a coarse-to-fine schedule to improve efficiency by performing more iterations at lower resolution in early stages and progressively increasing resolution for detail refinement, rather than operating at full resolution throughout. Moreover, the video generation model itself can be accelerated via few-step generation and attention-efficient designs.




\bibliography{dp}
\bibliographystyle{IEEEtran}

\end{document}